%% file: main.tex
\documentclass[sigconf]{acmart}

\AtBeginDocument{%
  \providecommand\BibTeX{{%
    \normalfont B\kern-0.5em{\scshape i\kern-0.25em b}\kern-0.8em\TeX}}}

\copyrightyear{2023}
\acmYear{2023}
\setcopyright{rightsretained}
\acmConference[FAccT '23]{2023 ACM Conference on Fairness, Accountability, and Transparency}{June 12--15, 2023}{Chicago, IL, USA}
\acmBooktitle{2023 ACM Conference on Fairness, Accountability, and Transparency (FAccT '23), June 12--15, 2023, Chicago, IL, USA}
\acmDOI{10.1145/3593013.3594116}
\acmISBN{979-8-4007-0192-4/23/06}

\usepackage{hyperref}
\usepackage{url}

\usepackage{graphicx}
\usepackage{algorithm}
\usepackage{algorithmic}
\usepackage{listings}

\usepackage{subfig}
\usepackage{enumitem}
\usepackage{tikz}
\usepackage{pgfplots}
\usepackage{rotating}
\usepackage{amsfonts}
\usepackage{multirow}
\usepackage[utf8]{inputenc}
\usepackage{booktabs}
\usepackage{nicefrac}
\usepackage{microtype}
\usepackage{xcolor}
\usepackage{bbold}
\usepackage{mathtools}

\usepgfplotslibrary{fillbetween}
\usetikzlibrary{patterns}
\usetikzlibrary{positioning}

\pgfplotsset{
    box plot/.style={
        /pgfplots/.cd,
        only marks,
        mark=-,
        mark size=1em,
        /pgfplots/error bars/.cd,
        y dir=plus,
        y explicit,
    },
    box plot box/.style={
        /pgfplots/error bars/draw error bar/.code 2 args={%
            \draw  ##1 -- ++(1em,0pt) |- ##2 -- ++(-1em,0pt) |- ##1 -- cycle;
        },
        /pgfplots/table/.cd,
        y index=2,
        y error expr={\thisrowno{3}-\thisrowno{2}},
        /pgfplots/box plot
    },
    box plot top whisker/.style={
        /pgfplots/error bars/draw error bar/.code 2 args={%
            \pgfkeysgetvalue{/pgfplots/error bars/error mark}%
            {\pgfplotserrorbarsmark}%
            \pgfkeysgetvalue{/pgfplots/error bars/error mark options}%
            {\pgfplotserrorbarsmarkopts}%
            \path ##1 -- ##2;
        },
        /pgfplots/table/.cd,
        y index=4,
        y error expr={\thisrowno{2}-\thisrowno{4}},
        /pgfplots/box plot
    },
    box plot bottom whisker/.style={
        /pgfplots/error bars/draw error bar/.code 2 args={%
            \pgfkeysgetvalue{/pgfplots/error bars/error mark}%
            {\pgfplotserrorbarsmark}%
            \pgfkeysgetvalue{/pgfplots/error bars/error mark options}%
            {\pgfplotserrorbarsmarkopts}%
            \path ##1 -- ##2;
        },
        /pgfplots/table/.cd,
        y index=5,
        y error expr={\thisrowno{3}-\thisrowno{5}},
        /pgfplots/box plot
    },
    box plot median/.style={
        /pgfplots/box plot
    }
}

\pgfplotsset{
	colormap={mygreen}{rgb255(0cm)=(254,254,254);rgb255(1cm)= (4,101,53)},
	colormap={myblue}{rgb255(0cm)=(160,30,50); rgb255(1cm)=(39,59,129)},
	colormap={myred}{rgb255(0cm)=(254,254,254); rgb255(1cm)=(160,30,50)},
	colormap={mybred}{rgb255(0cm)=(254,254,254); rgb255(1cm)=(217,10,100)},
	colormap={mybblue}{rgb255(0cm)=(217,10,100); rgb255(1cm)=(10,157,217)},
	colormap={mybgreen}{rgb255(0cm)=(254,254,254); rgb255(1cm)=(98,159,67)}
}

\definecolor{blue}{RGB}{31,120,180}
\definecolor{bblue}{RGB}{166,206,227}
\definecolor{green}{RGB}{ 251,154,153}
\definecolor{red}{RGB}{51,160,44}
\definecolor{bred}{RGB}{178,223,138}

\definecolor{rebuttal}{RGB}{255,0,0}

\input{macros}

\begin{document}

\title{On The Impact of Machine Learning Randomness on Group Fairness}

\author{Prakhar Ganesh}
\affiliation{%
  \institution{National University of Singapore}
  \country{Singapore}
}
\email{pganesh@comp.nus.edu.sg}
\author{Hongyan Chang}
\affiliation{%
  \institution{National University of Singapore}
  \country{Singapore}
}
\email{hongyan@comp.nus.edu.sg}
\author{Martin Strobel}
\affiliation{%
  \institution{National University of Singapore}
  \country{Singapore}
}
\email{martin.r.strobel@gmail.com}
\author{Reza Shokri}
\affiliation{%
  \institution{National University of Singapore}
  \country{Singapore}
}
\email{reza@comp.nus.edu.sg}

\input{sections/abstract}



\maketitle

\input{sections/sec_introduction}

\input{sections/sec_background}
\input{sections/sec_problem_statement}
\input{sections/sec_variance}
\input{sections/sec_uncertainty}
\input{sections/sec_data_order}
\input{sections/sec_applications}
\input{sections/sec_conclusion}

\begin{acks}
This research is supported by Google PDPO faculty research award, Intel within the www.private-ai.org center, Meta faculty research award, the NUS Early Career Research Award (NUS ECRA award number NUS ECRA FY19 P16), and the National Research Foundation, Singapore under its Strategic Capability Research Centres Funding Initiative. Any opinions, findings, conclusions, or recommendations expressed in this material are those of the author(s) and do not reflect the views of the National Research Foundation, Singapore.
\end{acks}

\bibliographystyle{ACM-Reference-Format}
\bibliography{references}

\section*{Image credits}
\begin{itemize}
\item Blueprint by Berkah Icon from \href{https://thenounproject.com/icon/blueprint-4093949/}{Noun Project}
\item Neural Network by Ian Rahmadi Kurniawan from \href{https://thenounproject.com/icon/neural-network-3875058/}{Noun Project}
\item Data by shashank singh from \href{https://thenounproject.com/icon/data-3235859/}{Noun Project}
    \item Data processing by Eko Purnomo from \href{https://thenounproject.com/icon/data-processing-5136443/}{Noun Project}
    \item Evaluate by Justin Blake from \href{https://thenounproject.com/icon/evaluate-4180849/}{Noun Project}
\end{itemize}

\clearpage
\appendix
\input{sections/sec_appendix_intro}
\input{sections/sec_appendix_datasets}
\input{sections/sec_appendix_hyperparameters}
\input{sections/sec_appendix_dropout}
\input{sections/sec_appendix_fairness_metrics}
\input{sections/sec_appendix_raw_runs}

\end{document}

%% file: macros.tex
\DeclareMathOperator*{\argmin}{argmin}

\newcommand{\Data}{\mathcal{D}}
\newcommand{\Algo}{\mathcal{A}}
\newcommand{\Loss}{\mathcal{L}}
\newcommand{\Functions}{\mathcal{F}}
\newcommand{\Indicator}{\mathbb{1}}
\newcommand{\Variance}{Var}

\newcommand{\martin}[1]{
	\if\compileFigures1
	\tikzexternaldisable
	\fi
	\todo[author=Martin,color=teal]{#1}
	\if\compileFigures1
	\tikzexternalenable
	\fi
}
\newcommand{\hongyan}[1]{
	\if\compileFigures1
	\tikzexternaldisable
	\fi
	\todo[author=Hongyan]{#1}
	\if\compileFigures1
	\tikzexternalenable
	\fi
}
\newcommand{\prakhar}[1]{
	\if\compileFigures1
	\tikzexternaldisable
	\fi
	\todo[author=Prakhar,color=cyan]{#1}
	\if\compileFigures1
	\tikzexternalenable
	\fi
}

%% file: sections/abstract.tex
\begin{abstract}
Statistical measures for group fairness in machine learning reflect the gap in performance of algorithms across different groups. These measures, however, exhibit a high variance between different training instances, which makes them unreliable for empirical evaluation of fairness. What causes this high variance? We investigate the impact on group fairness of different sources of randomness in training neural networks. We show that the variance in group fairness measures is rooted in the high volatility of the learning process on \emph{under-represented groups}. Further, we recognize the dominant source of randomness as the stochasticity of \emph{data order} during training. Based on these findings, we show how one can control group-level accuracy (i.e., model fairness), with high efficiency and negligible impact on the model's overall performance, by simply changing the data order for a single epoch.
\end{abstract}

%% file: sections/sec_introduction.tex
\section{Introduction}
\label{sec:introduction}

\begin{figure*}[hbtp]
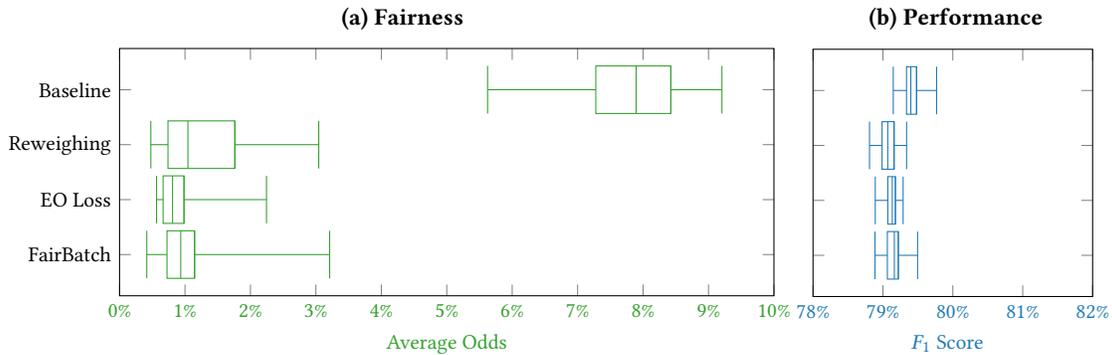

	\centering
	\addtolength{\tabcolsep}{-8pt}
	\begin{tabular}{cc}
	\textbf{(a) Fairness} & \textbf{(b) Performance} \\[0.2cm]	\input{figure_scripts/fig_variance_introduction_avgodds} & \input{figure_scripts/fig_variance_introduction_fscore} \\
	\end{tabular}
    \caption{ \textbf{Variance of fairness and accuracy:}
    \textbf{(a)} Fairness ({\color{red}average odds}) has a high variance across multiple runs due to non-determinism in training. This variance persists even with state-of-the-art bias mitigation algorithms (\textit{Reweighing~\citep{kamiran2012data}; Equalized Odds Loss~\citep{fukuchi2020fairtorch}; FairBatch~\citep{roh2020fairbatch}})
    \textbf{(b)} The overall performance ({\color{blue} $F_1$ score}), however, has a significantly smaller range of variance.}
	\label{fig:variance_introduction}
\end{figure*}

Machine learning models are shown to manifest and escalate historical biases present in their training data \citep{crawford2013hidden,barocas2016big,zhao2017men,abbasi2019fairness}. Understanding these biases and the resulting ethical obligations have led to the rise of fair machine learning research \citep{chouldechova2018frontiers,caton2020fairness,mehrabi2021survey}. 
However, recent work has observed high variance in fairness measures across multiple training runs, usually attributed to non-determinism in training (e.g., weight initialization, data reshuffling, etc.). These findings challenge the effectiveness of many bias mitigation algorithms \cite{amir2021impact,sellam2021multiberts}, and even the legitimacy of several fairness trends present in literature \cite{soares2022your}. Thus, a reliable extraction of fairness measures requires accounting for the high variance due to randomness in the learning process to avoid lottery winners (see Fig. \ref{fig:variance_introduction}).

The standard solution to this concern is executing a large number of training runs with different randomness. However, such a solution creates huge computational demands when examining biases in neural networks. For instance, it costs about $\$450K$ to train a model of similar quality as GPT-3 \cite{venigalla_li_2022}, and thus executing multiple training runs of such a model is not practical.
But, are multiple identical runs essential? Can we instead find an efficient alternative to measure this variance? Our paper answers this critical yet unsolved question. 
 
In this work, we perform an empirical investigation into the high fairness variance due to randomness in neural network training, with a diverse set of experiments on a multitude of settings, including different datasets across modalities, various fairness metrics, and changing hyperparameters and model architecture.
More specifically, our empirical analysis answers the following questions
\begin{itemize}[leftmargin=*]
    \item \textbf{Is there a dominant source of randomness?} We show that the fairness variance observed in the literature is dominated by randomness due to data reshuffling during training. Reshuffling causes large changes in fairness even between consecutive epochs within a single run, while other forms of randomness have minimal influence.
    \item \textbf{Why are fairness measures highly sensitive to data reshuffling?} We show a higher vulnerability of minorities to changing model behavior, i.e., a higher prediction uncertainty for under-represented groups. This disparate prediction uncertainty between groups is reflected in any statistical fairness measure defined on model predictions.
    \item \textbf{How does data order impact fairness?} We demonstrate an immediate impact of the data order on fairness. That is, we show that a model's fairness score is heavily influenced by the most recent gradient updates, irrespective of the preceding training. We also demonstrate how to create custom data orders that can efficiently control group-level performances (and thus in turn, model fairness), with a minor impact on the overall accuracy.
    \item \textbf{What are the practical implications?} 
    \begin{itemize}[leftmargin=*]
        \item Given the immediate impact of data order on model fairness and the nature of data order reshuffling in neural network training, we propose that using fairness variance across epochs in a \emph{single} training run is a good proxy to study fairness variance across multiple runs, thus reducing the computational requirements by a significant factor.
        \item We also propose a custom data order that can improve model fairness within a single training epoch, and compete with existing bias mitigation algorithms. Interestingly, we show that similar custom data orders can also be created by adversaries to \emph{freely control} fairness gaps in only a single epoch of training, even under explicit bias mitigation.
    \end{itemize}
\end{itemize}

%% file: sections/sec_background.tex
\section{Background and Related Work}
\label{sec:background}

In this section, we first introduce the relevant background in fair machine learning and randomness in neural network training. We then discuss the related work on the impact of randomness on fairness evaluation in deep learning.

\subsection{Fairness in Machine Learning}

Fair machine learning can be broadly divided into two categories, (i) group fairness \cite{chouldechova2018frontiers}, and (ii) individual fairness \cite{dwork2012fairness}. Group fairness relies on measuring the disparity between the average performance of protected groups against other privileged groups, and thus focuses on highlighting systematic bias against certain groups. 
Individual fairness instead relies on some form of similarity between individuals and requires consistency in the decision-making, i.e., similar individuals should be treated similarly.

In this work, we focus specifically on group fairness. Group fairness has a diverse set of definitions in the literature, usually chosen based on the stakeholders involved, known even to have opposing behavior in specific settings \cite{narayanan2018translation,saxena2019perceptions}. We will rely on three commonly used group fairness metrics, i.e., demographic parity, average odds, and equal opportunity~\cite{hardt2016equality}. Demographic parity is the measure of disparity between the percentage of positive outcomes for each group, i.e., it does not allow model predictions to depend on sensitive attributes. Average odds (and its relaxed version, equal opportunity) is instead a measure of disparity between predictions for each group conditioned on the true labels, i.e., it does allow overall predictions to depend on sensitive attributes, but does not allow predictions for certain ground-truth labels to depend on sensitive attributes.
Bias calculation and mitigation for group fairness have accumulated extensive literature in recent years, along with many open-source benchmarks \cite{reddy2021benchmarking,bellamy2018ai,bird2020fairlearn,fukuchi2020fairtorch}.


\subsection{Randomness in Neural Network Training}

Deep learning involves various forms of randomness that impact a neural network's path to convergence. This randomness during training can introduce noise into the optimization objective and works as a regularizer for the learning algorithm \cite{noh2017regularizing}. It makes the model prioritize generalization, avoid overfitting, escape local minima, and even speed up convergence \cite{bottou2012stochastic}.
Thus, randomness during training is integral to the success of neural networks, but its impact on model behavior needs to be carefully examined \cite{bethard2022we,black2022model}.

\begin{figure*}[hbtp]
	\centering
    \input{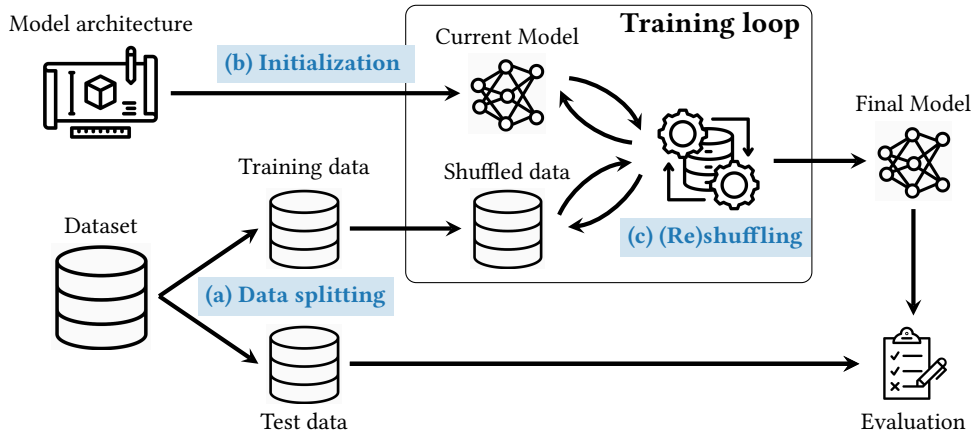} 
    \caption{\textbf{Sources of randomness during training:} Randomness is introduced at several points during the training of a neural network. \textbf{(a)} Data splitting before training is important to avoid information leakage during evaluation, and involves randomness for the same. \textbf{(b)} Randomness in weight initialization at the start of the training is necessary to break symmetry and learn complex representations. \textbf{(c)} Random reshuffling at every epoch implements stochastic gradient descent (SGD) in practice. There are many other forms of randomness in neural network training. Yet, most are only used for certain specific settings.
    }
	\label{fig:overview}
\end{figure*}


Broadly, randomness in neural networks can be studied in the context of the following categories (see Fig. \ref{fig:overview}),

\begin{itemize}[leftmargin=*]
    \item \textbf{Data Splitting:} For any experimental setup in machine learning, the dataset under consideration is randomly divided into train-val-test (or just train-test) splits, to avoid information leakage and perform a fair evaluation.
    \item \textbf{Weight Initialization:} 
    Weight initialization refers to the initial parameter vector that is the starting point for the gradient descent. Randomness in weight initialization is crucial for breaking the symmetry between model parameters and allows the neural network to learn complex functions \cite{ghatak2019initialization}. 
    \item \textbf{Random Reshuffling:} Neural network training relies on gradient descent to optimize a chosen objective iteratively. Calculating the gradient over the entire dataset for every optimization step is expensive. A commonly adopted alternative is uniformly sampling a subset of the dataset to approximate the gradient, known as stochastic gradient descent (SGD). In practice, it has been shown that instead of uniform sampling, SGD can also be implemented by simply traversing a random order, i.e., random reshuffling, of the training data \cite{bottou2012stochastic,mishchenko2020random,nguyen2021unified}.
    \item \textbf{Other Sources of Randomness:} Several additional components in the learning algorithm can introduce further randomness. These components are not standard, but are introduced in special cases to achieve specific objectives. For example, data augmentation to increase dataset size \cite{khosla2020enhancing}, dropout to perform regularization \cite{wager2013dropout}, gradient noise for private training \cite{chaudhuri2011differentially}, etc.
\end{itemize}

\subsection{High Variance in Fair Deep Learning}
There has been a growing awareness of high variance in fair deep learning, associated with non-determinism in model training or the underlying implementation \cite{amir2021impact,sellam2021multiberts,soares2022your,qian2021my,friedler2019comparative}, and the uncertainty of existing results in the literature.

Soares et al. \cite{soares2022your} investigate the relationship between various algorithmic choices and the corresponding fairness variance, in large language models. They study the correlation of fairness with model size and found no obvious trends, as opposed to various claims previously made in literature \cite{bender2021dangers,hooker2020characterising}. They also found that fairness is heavily affected by the random seed, i.e. simply changing the randomness can cause a huge variance in fairness.

Sellam et al. \cite{sellam2021multiberts} have trained and released 25 pre-trained BERT checkpoints, each trained from scratch under identical settings but with a different random seed. They also analyze the variance of model fairness and the impact of commonly used bias mitigation algorithms on downstream tasks when starting with different pre-trained models.
They show significant variance across changing random seeds and question the value of such mitigation techniques.

Amir et al. \cite{amir2021impact} revisit bias mitigation techniques in clinical texts and show a lack of statistically significant improvement after accounting for non-determinism in training. 
Friedler et al. \cite{friedler2019comparative} explore the stability of fairness under a rarely studied source of randomness, i.e. data splitting, and show notable impact on fairness evaluation.

While existing literature focuses on exploring the impact of high fairness variance in bias evaluation, we instead focus on investigating its source. Furthermore, we propose to move away from the practice of simply executing multiple runs to capture fairness variance and instead provide a computationally efficient proxy.

%% file: sections/sec_problem_statement.tex
\section{Problem Statement}
\label{sec:problem}

We start by formally defining the problem statement and detailing our experiment setting for the rest of the paper.

\subsection{Neural Network Training} 
Most machine learning algorithms can be abstracted down to an optimization problem for a given objective, usually a loss function. More specifically, for a training dataset $(x, y) \in \Data$, a family of hypothesis functions $\Functions$, and a loss function $\Loss$, the optimization goal for the learning algorithm can be defined as,
\begin{align}
    f^* \leftarrow \argmin\limits_{f \in \Functions} \sum\limits_{(x, y) \in \Data} \Loss(f(x), y) 
\end{align}

The above formulation of the learning objective is known as empirical risk minimization (ERM) \cite{vapnik1991principles}. However, finding a global optimum for ERM in deep learning is typically intractable, due to the high dimensional, non-convex formulation of neural networks. Neural networks are instead trained iteratively, starting with a randomly sampled function $f_0$, refining the model with a learning algorithm $\Algo$ for $T$ epochs, to finally output the trained model $f_T$. The learning algorithm at every epoch $t$ takes in the current model, complete training data $\Data$, and a number of hyperparameters $\xi$ (e.g., batch size, learning rate, etc.), to progressively improve the model by a single epoch of training. The learning algorithm can contain various sources of randomness, as discussed above. In our work, we will focus on two standard forms of randomness found in every neural network training, i.e., weight initialization and random reshuffling of data order at every epoch. More specifically, neural network training can be defined as,
\begin{align}
    f_t := \Algo(f_{t-1}, \Data, \xi, r_s, t) \quad \quad 
    f_0 \sim \Functions; r_s \sim R
\end{align}

The function $f_0$, i.e., the weight vector initialization in a parameterized neural network, is randomly sampled from pre-defined distribution $\Functions$, and the random seed for reshuffling $r_s$ is sampled from a uniform distribution $R$.
Note that both random seed $r_s$ and epoch number $t$ are together responsible for the data shuffling of epoch $t$. Thus, for fixed reshuffling (i.e., fixed $r_s$), the data order is still shuffled at every epoch during a single training run but is the same at any epoch $t$ across two different training runs.
More details on the random seed setup can be found in Appendix A.

\subsection{Metrics and Variance} 
A model $f$'s performance can be evaluated using its outputs on the test dataset. We will stick to the commonly used binary classification and binary sensitive attribute $a \in \{0, 1\}$ setting in fairness literature for the rest of the discussion. In the main paper, we will rely on $F_1$ Score and average odds (AO) to measure the model's average performance and group fairness respectively. AO can be empirically interpreted as the average disparity between separately calculated \textit{true positive rates} (TPR) and \textit{false positive rates} (FPR) of various groups \cite{hardt2016equality}. The metrics are defined as
\begin{flalign}
    & F_1(f,\Data) := \frac{2TP}{P + PP} = \frac{2 \sum_{\Data^e} \Indicator_{[f(x)=y \wedge y=1]}}{\sum_{\Data^e} \Indicator_{[y=1]} + \sum_{\Data^e} \Indicator_{[f(x)=1]}} \\
    & AO(f,\Data) := \frac{(\Delta TPR + \Delta FPR)}{2} \\
    & = \frac{1}{2} \sum\limits_{\substack{h=\{0,1\}\\y=h}} |\frac{\sum_{\Data^e} \Indicator_{[f(x)=1 \wedge a=0]}}{\sum_{\Data^e} \Indicator_{[a=0]}} - \frac{\sum_{\Data^e} \Indicator_{[f(x)=1 \wedge a=1]}}{\sum_{\Data^e} \Indicator_{[a=1]}}| \notag
\end{flalign}

where $\sum_{\Data^e}$ is the sum over all data points in the test set $\Data^e$, i.e., $\sum_{(x, y, a) \in \Data^e}$, and $\Indicator_{[z]}$ is an indicator function which is $1$ when the boolean expression $z$ is true, and $0$ otherwise.
We also include additional experiments for two more fairness measures, equal opportunity (EOpp) and demographic parity (DP) in Appendix~I.
Moreover, we will show that the non-determinism in fairness originates from high prediction uncertainty for minority (Section~\ref{sec:group_variance}), and thus will be reflected in any fairness metric defined on these predictions. We report all metrics in percentage.

At the heart of our work is the study of fairness variance across model checkpoints. We define variance across multiple runs and variance across epochs in a single run as,
\begin{align}
    \Variance^{runs}_{F_1}(\Algo, T) := \underset{f_0 \sim \Functions; r_s \sim R}{\Variance}(F_1(f_T)), \\
    \Variance^{epochs}_{F_1}(\Algo, f_0, r_s, T_1, T_2) := \underset{t \in [T_1, T_2]}{\Variance}(F_1(f_t)), \\
    \Variance^{runs}_{AO}(\Algo, T) := \underset{f_0 \sim \Functions; r_s \sim R}{\Variance}(AO(f_T)), \\ 
    \Variance^{epochs}_{AO}(\Algo, f_0, r_s, T_1, T_2) := \underset{t \in [T_1, T_2]}{\Variance}(AO(f_t)),
\end{align}

Existing work in the literature has shown high variance in fairness scores across multiple runs $\Variance^{runs}_{AO}(\Algo, T)$. In our work, we first decouple the impact of two standard sources of randomness, i.e., study $\underset{f_0 \sim \Functions}{\Variance}(F_1(f_T))$ and $\underset{r_s \sim R}{\Variance}(F_1(f_T))$ separately. In doing so, we find high variance in fairness scores even across epochs in a single training run (Section \ref{sec:variance}), and thus further study variance across epochs in fairness scores $\Variance^{epochs}_{AO}(\Algo, f_0, r_s, T_1, T_2)$. Note that for $F_1$ score, variance across multiple runs $\Variance^{runs}_{F_1}(\Algo, T)$ and across epochs in a single run $\Variance^{epochs}_{F_1} (\Algo, f_0, r_s, T_1, T_2)$ are both relatively stable. Unless otherwise specified, we train our models for a total of $T=300$ epochs, and we measure variance across epochs from $T_1=100$ to $T_2=300$.  We make this choice because the models have converged to stable accuracy before epoch $100$ (refer to the training curve in Appendix C for more details).

\subsection{Datasets and Models} 
We will conduct our investigation on ACSIncome and ACSEmployment tasks of the Folktables dataset \citep{ding2021retiring}, and binary classification of the 'smiling' label in CelebA dataset \citep{liu2015faceattributes}, with perceived gender (Female vs. Male) as the sensitive attribute for all datasets. For CelebA, input features are obtained by passing the image through a pre-trained ResNet-50 backbone and extracting the output feature vector. More details on the datasets are provided in Appendix B.

We train a feed-forward network with a single hidden layer of $64$ neurons and \textit{ReLU} activation, and train the model with \textit{cross-entropy (CE)} loss for $T=300$ epochs at batch size $128$ and learning rate $1e-3$, in all our experiments unless specified otherwise. Note that while we measure fairness scores in our experiments, we do not explicitly train the models with any fairness constraints (except in Section \ref{subsec:manipulation_mitigation} when training baseline bias mitigation algorithms). We also include additional experiments by changing training hyperparameters, i.e., batch size, learning rate, and model architecture, in Section~\ref{subsec:correlation_others} (and Appendix~G).
We use a train-val-test split of $0.7:0.1:0.2$, and maintain the same split throughout all our experiments, i.e. we do not consider potential randomness due to data splitting. All our experiments and evaluations are performed only on the test split. We will focus primarily on the ACSIncome task in the main text, while additional experiments on CelebA and the ACSEmployment task are included in Appendix F.

%% file: sections/sec_variance.tex
\section{The Dominant Source of Randomness}
\label{sec:variance}

In this section, we move past the observation that different training runs lead to different outcomes, and investigate the high fairness variance by studying and contrasting the two canonical sources of randomness.

\subsection{Impact of Weight Initialization and Random Reshuffling on Fairness Variance}
\label{subsec:weight_shuffling}

\begin{figure*}[htbp]
	\centering
    \input{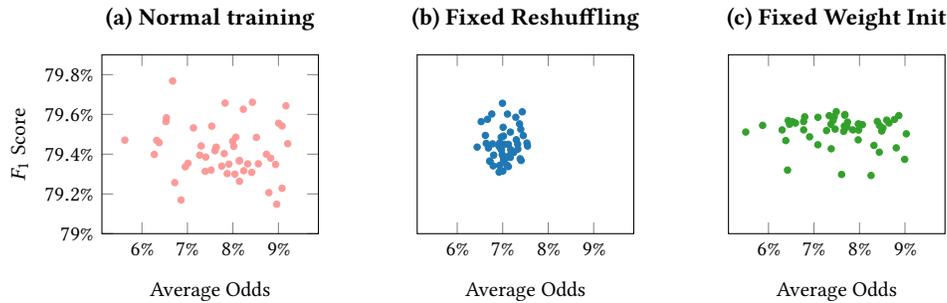}
    \caption{
    \textbf{Decoupling the effect of randomness in weight initialization and reshuffling:} 
    \textbf{(a)} Variance in average odds (AO) by allowing both sources of randomness simultaneously represents the fairness variance in existing literature. \textbf{(b)} We see a significant drop in variance if we change only the weight initialization while keeping the reshuffling fixed. \textbf{(c)} However, we observe high range of variance by changing only the reshuffling, even for a fixed weight initialization. These results suggest reshuffling of the data order as the dominant source of fairness variance, with little influence from weight initialization.}
	\label{fig:decouple_scatter}
\end{figure*}

\begin{figure*}[htbp]
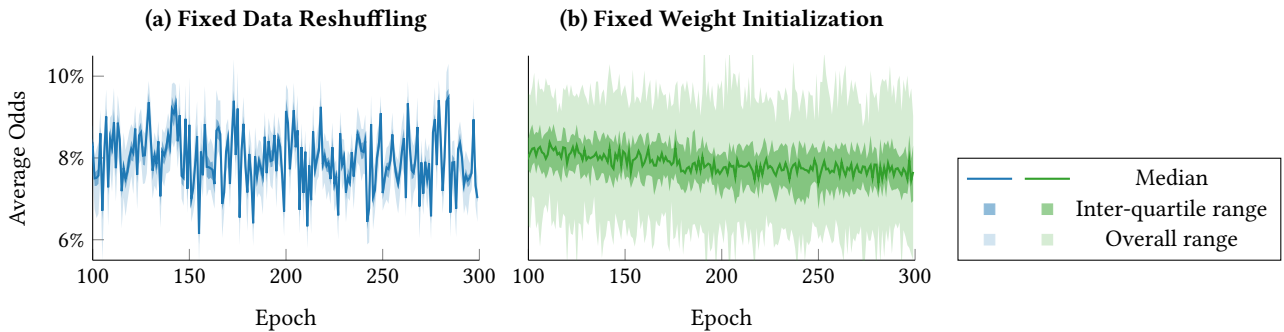

	\centering	\input{figure_scripts/fig_curve_correlation}
    \input{figure_scripts/fig_curve_correlation_legend}
    \caption{\textbf{Training dynamics under fixed weight initialization and reshuffling:} Median, inter-quartile range, and overall range of average odds across 50 training runs while keeping the data reshuffling or the weight initialization fixed respectively. \textbf{(a)} Despite different initializations, models with fixed data reshuffling have very little variance across training runs, but high variance across epochs. This highlights the dominant impact of random reshuffling on model fairness. \textbf{(b)} High variance even across training runs at the same epoch under fixed weight initialization further supports our claim.
    }
	\label{fig:decouple_correlation}
\end{figure*}

We start by decoupling the two sources of randomness inherent to the widely adapted SGD, i.e., weight initialization and random reshuffling, and study their impact on fairness variance separately in Fig. \ref{fig:decouple_scatter}. We collect average odds (AO) and $F_1$ score at epoch $300$ for 50 unique training runs each while, \textit{(i)} allowing for both sources of randomness, \textit{(ii)} changing only the weight initialization while keeping the random reshuffling fixed, and \textit{(iii)} changing only the random reshuffling while keeping the weight initialization fixed, respectively.
The large range of fairness scores reported by allowing for both sources of randomness in Fig. \ref{fig:decouple_scatter}(a) represents the variance observed in the existing literature. 
Interestingly, when these sources are examined separately, the variance under fixed initialization in Fig. \ref{fig:decouple_scatter}(c) is equivalently large but the variance under fixed reshuffling in Fig. \ref{fig:decouple_scatter}(b) drops significantly. It is clear that fairness variance originates from data order shuffling, while randomness in weight initialization has minimal impact.


To further probe the difference between these two sources of randomness, we study the training dynamics of the previous set of models across epochs, instead of just the final model checkpoint, in Fig. \ref{fig:decouple_correlation}. We plot the median, inter-quartile range, and overall range of average odds (AO) across the complete set of 50 training runs from epoch $100$ to $300$ in the two isolated settings from the previous experiment.
We find a high correlation in fairness scores across training runs with fixed data reshuffling (average pairwise pearson coefficient $\approx 0.94$), which supports our observations of low variance in fairness scores at the final epoch in Fig. \ref{fig:decouple_scatter}(b).
Furthermore, there is a lack of any reasonable correlation between fairness scores of training runs with fixed weight initialization (average pairwise pearson coefficient $\approx 0.04$).
Interestingly, the high fairness variance across epochs inside a single training run in Fig. \ref{fig:decouple_correlation}(a) closely matches the variance that we observe across multiple training runs in Fig. \ref{fig:decouple_correlation}(b). 
In other words, for fixed data reshuffling the average odds value at any epoch is almost the same between different training runs, but the average change even between consecutive epochs is large, while for fixed weight initialization, even the variance between runs is quite high.

\subsection{Dominance of Random Reshuffling across Datasets, Metrics and Hyperparameters}
\label{subsec:correlation_others}

\begin{table*}
    \begin{center}
    \caption{\textbf{Average pairwise pearson coefficient for correlation across multiple runs:} Fixed random reshuffling (RR) (i.e., changing only the weight initialization) has a high correlation score across multiple runs, while fixed weight initialization (WI) (i.e., changing only the random reshuffling) has a correlation score close to zero, which establishes the dominance of data reshuffling on fairness. These trends exist across different datasets, fairness metrics, and hyperparameter choices.}
    \label{tab:correlation_others}
    \begin{minipage}{0.48\textwidth}
        \centering
        \textbf{(a) Different Datasets}\\[0.2cm]
        \begin{tabular}{lcc}
        \toprule
         & Fixed RR & Fixed WI \\
        \midrule
        ACSIncome & .94 & .04 \\
        ACSEmployment {\color{white}aa} & .89 & .01 \\
        CelebA & .92 & .01 \\
        \bottomrule
        \end{tabular} \\ [0.2cm]
        \textbf{(b) Different Fairness Metrics}\\[0.2cm]
        \begin{tabular}{lcc}
        \toprule
         & Fixed RR & Fixed WI \\
        \midrule
        Average Odds & .94 & .04 \\
        Equal Opportunity & .94 & .05 \\
        Demographic Parity & .96 & .03 \\
        \bottomrule
        \end{tabular} \\
    \end{minipage}\hfill
    \begin{minipage}{0.48\textwidth}
        \centering
        \textbf{(c) Changing Hyperparameters}\\[0.2cm]
        \begin{tabular}{lcc}
        \toprule
         & Fixed RR & Fixed WI \\
        \midrule
        Default Hyperparameters & .94 & .04 \\
        Batch Size = 16 & .95 & .00 \\
        Learning Rate = 0.01 & .93 & .00 \\
        Arch = \{2048, 64\} & .92 & .00 \\
        \addlinespace[2ex]
        No Dropout & .94 & .04 \\
        Dropout Rate = 10\% & .88 & .03 \\
        Dropout Rate = 20\% & .85 & .04 \\
        Dropout Rate = 30\% & .80 & .13 \\
        \bottomrule
        \end{tabular} \\ [0.2cm]
    \end{minipage}\hfill
    \end{center}
\end{table*}

We extend our previous experiment and calculate correlation across multiple runs for additional datasets, fairness measures, as well as hyperparameter choices of batch size, learning rate, model architecture, and dropout regularization with different dropout rates in Table \ref{tab:correlation_others}. 
Here we measure the correlation (i.e., average pairwise pearson coefficient) across 50 training runs in each setting for fixed data shuffling and fixed weight initialization.
It is clear from the results that even under diverse settings, the correlation between multiple runs with fixed data reshuffling is significantly high, while the correlation with fixed weight initialization is close to zero. 



In addition to the overall trends supporting our initial claim, individual trends in Table \ref{tab:correlation_others} under various settings are also quite interesting. The correlation score for fixed weight initialization under hyperparameters that induce noisier training (i.e., smaller batch size, larger learning rate, etc.) drops even lower than the default setup. Note that we do not report results for a bigger batch size or a smaller learning rate, as these models face issues with convergence even after $300$ epoch of training (see Appendix G for more details), which further indicates the need of randomness and noise in the learning algorithm to facilitate better and faster convergence. However, this randomness will also create a dominant dependence of model fairness on data reshuffling, as we study in our work. We also see the trends diminishing (although still clear) with higher dropout rates. This suggests that appropriate regularization during training can indeed, to some extent, reduce the impact of randomness in training (or more specifically, data reshuffling) on fairness scores.
We also provide detailed results for each hyperparameter setting above in Appendix G, H.

\textit{\textbf{Takeaway 1:} Random reshuffling of data order during training is the dominant cause of high fairness variance as seen in the literature, while randomness in weight initialization has minimal influence.}

%% file: sections/sec_uncertainty.tex
\section{Why is Fairness Highly Sensitive to Randomness?}
\label{sec:group_variance}

In the previous section, we showed the dominant impact of data reshuffling on model fairness. In this section, we show that its in fact the imbalance in the underlying data distribution for training which creates high volatility in predictions for the minority. Thus, groups with smaller representations are more significantly influenced by the randomness in reshuffling, resulting in high fairness variance.

\subsection{Changing Predictions Across Epochs}

We observed high variance in model fairness even between consecutive epochs during training (see Fig. \ref{fig:decouple_correlation}). The changing predictive behavior of neural networks beyond training loss convergence is not surprising, and has been studied extensively in literature \citep{toneva2018empirical,kirkpatrick2017overcoming,jagielski2022measuring,tirumala2022memorization}. As we are concerned with the fairness of the final decisions made by the model, we will focus on a change in the model's discrete output class when discussing changing predictions. More specifically, a model is said to have undergone a change in prediction for some input $x$ during epoch $t$, if $f_t(x) \neq f_{t-1}(x)$, where $f_t(x)$ is the output class when passing the input $x$ through the model checkpoint at the end of epoch $t$.
While these changing predictions maintain an overall stable average performance, they can still have a disparate impact on individual groups, the exact characteristics of which are less known.

\begin{figure*}[hbtp]
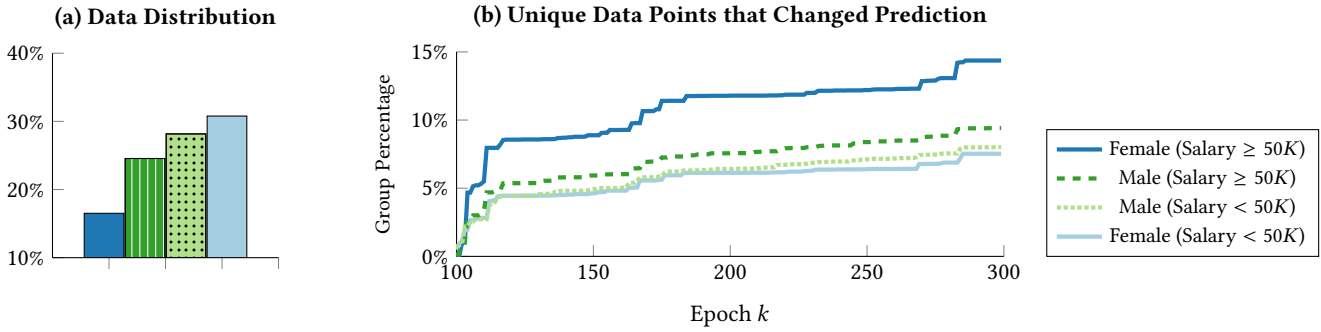

	\centering
    \input{figure_scripts/fig_data_distrib}
    \input{figure_scripts/fig_example_forgetting_total}
    \raisebox{1.0357cm}{
    \input{figure_scripts/fig_example_forgetting_legend}
    }
    \caption{\textbf{Disparate percentage of changing predictions across epochs:} \textbf{(a)} The underlying data distribution of ACSIncome shows positive labels from group Female as an under-represented minority. \textbf{(b)} Total percentage of unique data points from each subgroup that change prediction across epochs follow the opposite order of their representation in the training data. These results highlight subgroups with least representation being the most vulnerable to changing predictions.}
	\label{fig:example_forgetting}
\end{figure*}

\begin{figure*}[htbp]
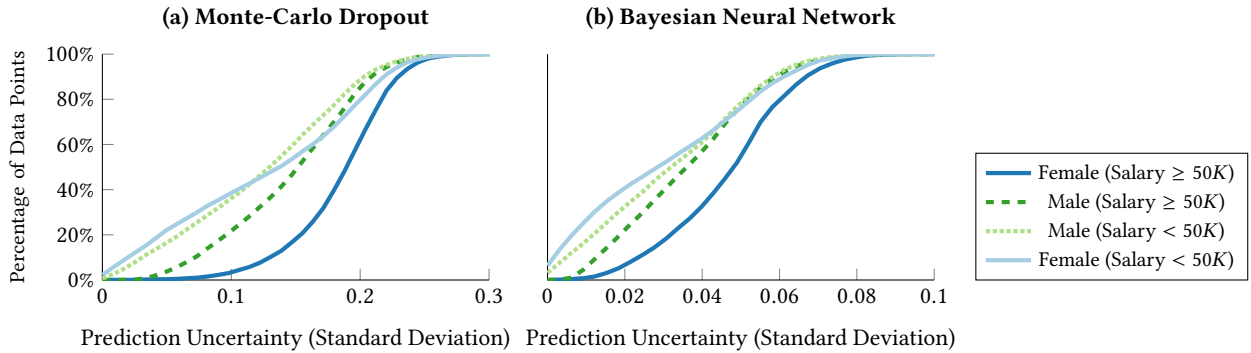

	\centering
	\input{figure_scripts/fig_uncertainty.tex}
    \raisebox{1.0342cm}{
    \input{figure_scripts/fig_example_forgetting_legend}
    }
    \caption{\textbf{Normalized cumulative distribution of prediction uncertainty for various groups:} The distribution of the minority group ({\color{blue} \textbf{Female with Salary $\geq 50K$}}) is significantly more skewed towards higher uncertainty than any other group, i.e. the minority contains far more percentage of data points with high prediction uncertainty than the majority.
    }
	\label{fig:uncertainty}
\end{figure*}

We study this instability by investigating individual data points which change their predictions. We plot the dataset distribution across groups in Fig. \ref{fig:example_forgetting}(a) and the percentage of data points from each group that changed their prediction at least once between epochs $100$ and $k$, where we gradually increase the value of $k$, in Fig. \ref{fig:example_forgetting}(b).
Clearly, the trends in the percentage of unique data points with changing predictions mirror the representation of each group in the original dataset, i.e. the groups which are represented the least are the most vulnerable to changing model behavior. For example, positive labels from the group Female are severely under-represented and consequently have almost twice the percentage of unique examples with changing predictions than any other groups.

\subsection{Disparate Prediction Uncertainty}
Higher vulnerability to changing discrete predictions for minorities can be interpreted as an indication of higher uncertainty in the underlying model predictions. To further probe the disparate model behavior across different groups, we record the cumulative distribution of prediction uncertainty for each group separately in Fig. \ref{fig:uncertainty}. We rely on two commonly used methods to measure prediction uncertainty, i.e., \textit{(i)} monte-carlo dropout \citep{gal2016dropout}, and \textit{(ii)} training a bayesian neural network \cite{blundell2015weight}. We execute 1000 forward passes for each method and record the standard deviation in outputs as the prediction uncertainty.
Despite different distribution trends, both methods highlight the higher prediction uncertainty of minorities.
As expected, the order of prediction uncertainty across groups follows the training data distribution (Fig. \ref{fig:example_forgetting}(a)), i.e. groups with a larger representation in the training data have smaller number of examples with large prediction uncertainty, which is quite intuitive. As the model has skewed cumulative distribution towards uncertainty for the minority, any fairness metric defined on the output of such a model will also reflect this instability, and thus manifests as fairness variance in existing literature \citep{amir2021impact,sellam2021multiberts,soares2022your}.

\textit{\textbf{Takeaway 2:} Under-represnted groups have higher prediction uncertainty in the final trained model, and thus predictions for data points from such minorities are more sensitive to the randomness.}

%% file: sections/sec_data_order.tex
\section{Impact of Data Order on Fairness}
\label{sec:data_order}

\begin{figure*}[htbp]
    \centering
    \input{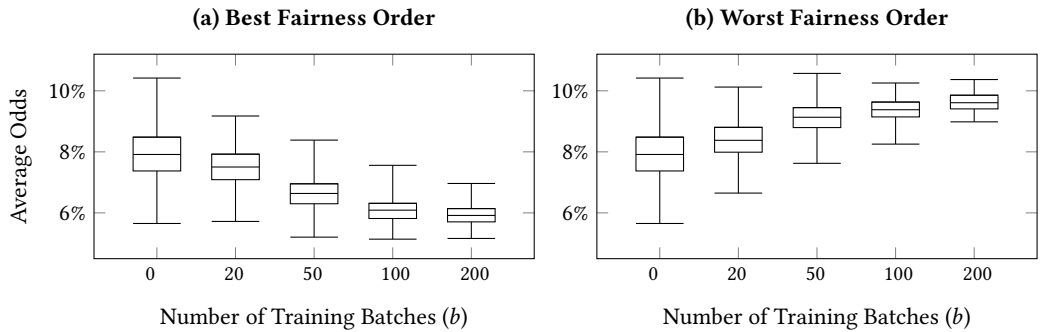}
    \caption{\textbf{Fairness variance under common $b$ most recent gradients:} Average odds stabilize as the number of most recent common training batches $b$ increases, highlighting the immediate impact of these gradient updates on model fairness. Moreover, it even predictably stabilizes to low or high fairness based on the corresponding data order from the epoch with best or worst fairness.}
    \label{fig:suffix_shuffle}
\end{figure*}

\begin{figure*}[htbp]
	\centering
    \input{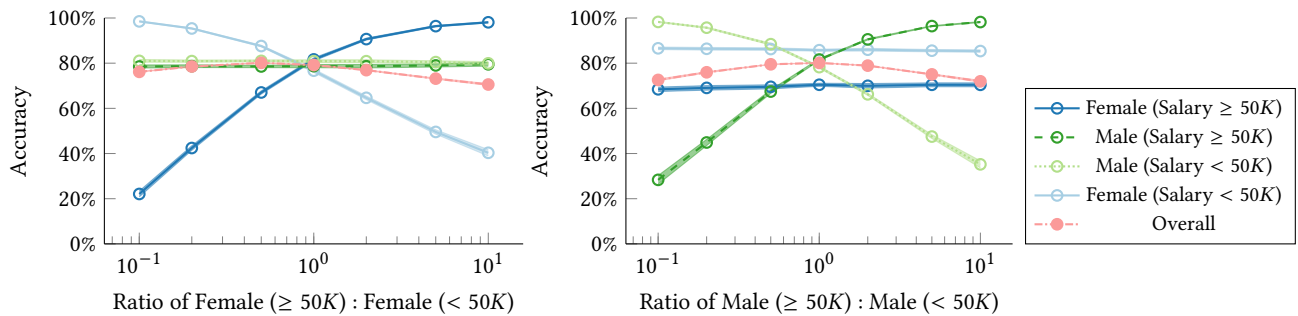} 
    \caption{\textbf{Manipulating group level accuracy with data order:} We show how to control group level accuracy by changing data distribution of the most recent gradient updates, tested separately for ratio between positive and negative labels for group Female, and group Male, respectively, while keeping other ratios fixed. In only a single epoch of training, we are able to manipulate the group level accuracy trade-off, with relatively small impact on overall accuracy.
    }
	\label{fig:manipulate}
\end{figure*}

In Section \ref{sec:variance}, we observed the dominance of reshuffling on fairness variance. Data order during training governs gradient updates and thus its impact on fairness is unsurprising \citep{shah2020choosing,soviany2022curriculum,mohtashami2022characterizing,shumailov2021manipulating,rajput2021permutation,lu2022grab}. Even under randomly shuffled data order, neural networks are known to undergo changes in predictive behavior during training \cite{toneva2018empirical,kirkpatrick2017overcoming,jagielski2022measuring,tirumala2022memorization}. However, its the immediacy of the impact of data order that is surprising and a novel observation of our work. 
We now study the impact of data order in a single epoch on fairness.

\subsection{Data Order's Immediate Impact}

To study the immediacy and characteristics of the impact of data order on model fairness, we fine-tune a set of already converged model checkpoints for a common sequence of $b$ batches and record the fairness variance across checkpoints for different values of $b$ in Fig. \ref{fig:suffix_shuffle}. This allows us to measure fairness variance across models which have experienced the same most recent $b$ gradient updates. The choice of the common sequence of $b$ batches fed to the model was done by separately training a model and choosing the suffix of data order corresponding to epochs with the best and worst fairness scores on the validation set. As the number of fixed batches $b$ increases, the fairness variance decreases, and it is clear from the results that the impact of data order on fairness is quite immediate (an epoch of ACSIncome dataset is $b=1070$ batches). Moreover, the resulting fairness is also characteristically stable for a specific data order, i.e. batches taken from the suffix of the data order corresponding to the best fairness epoch of an individual training run also helps fine-tune all other checkpoints towards the same best fairness, and vice-versa for the worst fairness epoch.

The set of checkpoints were chosen by sampling 1000 different checkpoints from epochs $100$ to $300$ of 50 different training runs while allowing for both forms of randomness simultaneously. The checkpoints were chosen in this manner to create a diverse training history, and show that these models achieve the same fairness scores when fine-tuned on a common set of batches, irrespective of training prior to those updates.
We also extend our experiment to show the same behavior even for batches from any random data order, instead of deliberately chosen data orders as above, in Appendix D.
One possible explanation of this immediate impact is the presence of no energy valleys in deep learning loss landscape between minima of separately trained models \citep{draxler2018essentially,garipov2018loss}. Another possible explanation can be built on a recent line of work that shows there is only one functionally unique minima in loss landscape of neural networks, while all other minima simply contain permutation or scale symmetries of the same set of models \citep{ainsworth2022git}.

\subsection{Manipulating Group Accuracy Distribution with Data Order}
\label{subsec:manipulating}

In the previous section, we saw a stable relationship between data order and the resulting fairness score. However, the data orders in the experiment above were sampled from a set of random data orders. We now show that it is possible to create our own custom data order to achieve any target fairness score. We hypothesize that since the data order in the most recent batches has an immediate impact on model fairness, the distribution in these batches must be temporarily changing the loss landscape and nudging the group-level accuracy. This could allow us to manipulate group-level accuracy in only a single epoch of fine-tuning.

\begin{figure*}[hbtp]
    \centering

    \begin{minipage}{0.48\textwidth}
        \centering
        \textbf{(a) Fairness Score Distribution}\\[0.2cm]
        \includegraphics[width=0.9\linewidth]{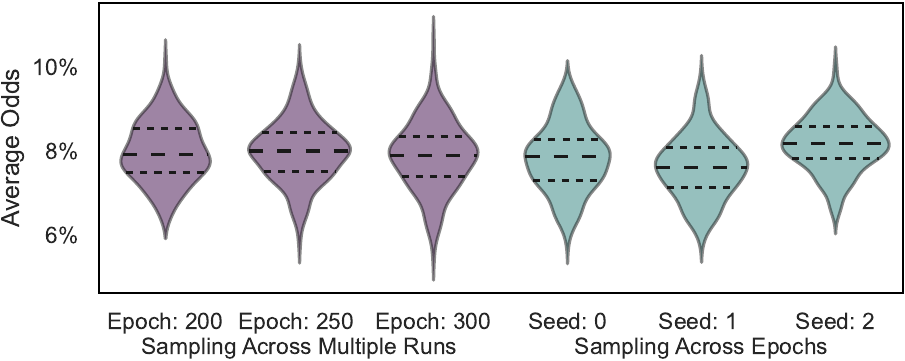}
    \end{minipage}\hfill
    \begin{minipage}{0.48\textwidth}
        \centering
        \textbf{(b) Black Swans} \\[0.2cm]
        \begin{minipage}{0.5\textwidth}
        \includegraphics[width=0.95\linewidth]{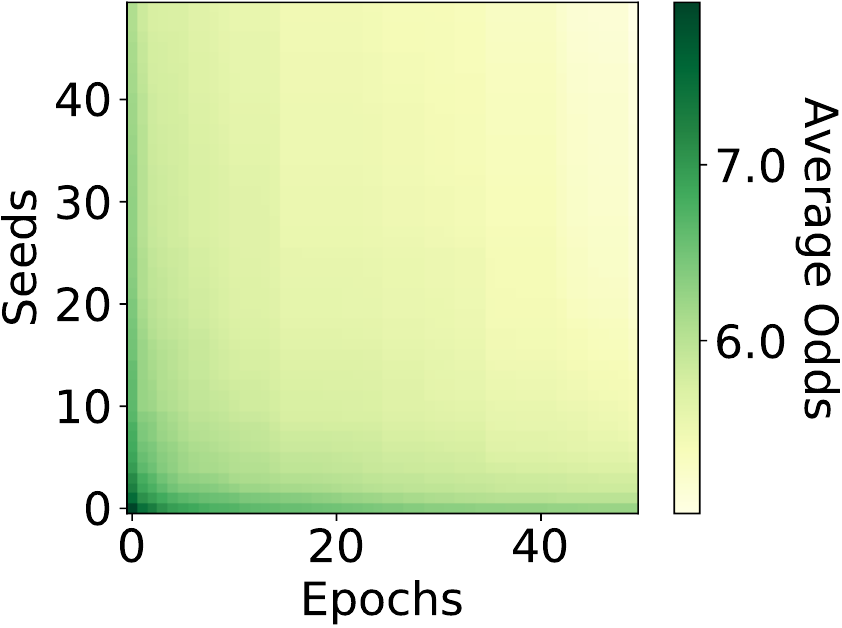}
        \end{minipage}\hfill
        \begin{minipage}{0.5\textwidth}
        \includegraphics[width=0.95\linewidth]{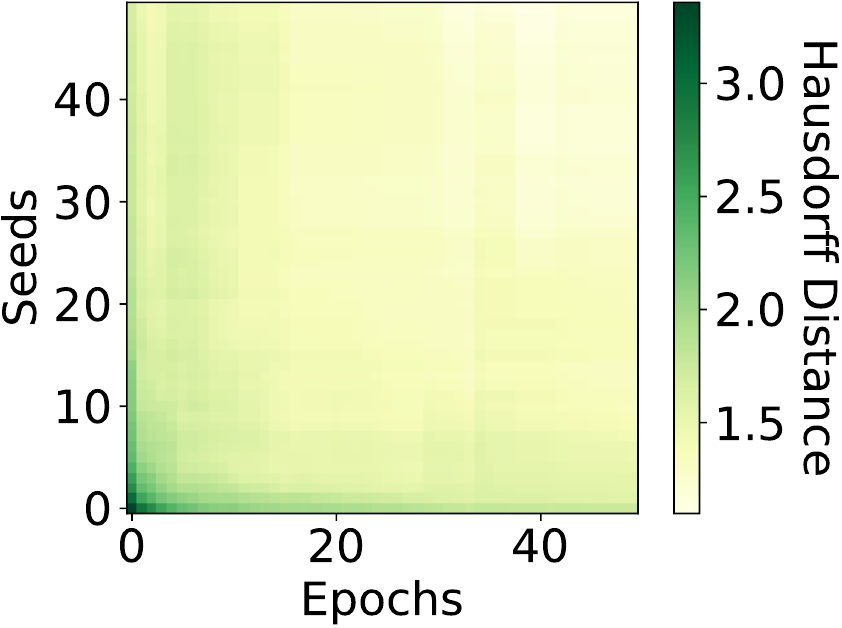}
        \end{minipage}\hfill
    \end{minipage}\hfill
    \caption{\textbf{Fairness variance across multiple runs vs across epochs in a single run:} \textbf{(a)} Fairness scores (average odds) across multiple training runs and across epochs in a single training run have similar empirical distributions. \textbf{(b)} Quality of black swans (i.e. extremely rare checkpoints) improves with more checkpoints collected, either in terms of fairness (the lowest achievable average odds score) or the trade-off between overall performance and fairness (the Hausdorff distance to the best pareto front). This improvement occurs at the same rate, irrespective of sampling checkpoints across multiple random seeds (x-axis) or multiple epochs in a single seed (y-axis). But, sampling across epochs is significantly cheaper, providing a highly efficient alternative to executing multiple runs.}
    \label{fig:variance_single_run}
\end{figure*}

To test this hypothesis, we fine-tune a set of 50 already converged models for exactly a single epoch on custom data orders with chosen distributions and record group-level accuracy in each setting in Fig. \ref{fig:manipulate}.
To create this custom data order, we start by fixing the ratio between different groups and then form batches for the data order suffix in this exact ratio until we run out of data points (which will happen for any ratio that is not the dataset distribution). The excess data points are then shuffled randomly and placed at the prefix of data order. We do two sets of isolated experiments, by changing the ratio between positive and negative labels for group Female (Fig. \ref{fig:manipulate}(a)) and group Male (Fig. \ref{fig:manipulate}(b)) respectively, while keeping other ratios fixed.
It is clear that by manipulating the data distribution in the most recent gradient updates, we can control the group-level accuracy of the model. While the overall accuracy drops noticeably for extreme ratios, it does not change much in the middle despite significant variance in group-level performances.

Note that our custom data order still has the same distribution as the original dataset, i.e., we have not changed the distribution of the complete data order, but only moved around the data points to change the distribution of the data order suffix.
These results further strengthen our claim on the immediate impact of the most recent gradient updates on model fairness and group-level accuracy. \textit{Fairbatch} \citep{roh2020fairbatch}, a recently proposed bias mitigation algorithm, follows a similar formulation (although it additionally changes the overall distribution). Fairbatch creates batches with a fixed ratio between groups, and this ratio is continuously optimized to counter the existing bias in the model. Our results not only explain their success, but also state that instead of regularly adapting the distribution to compensate for the model bias (and oversampling/undersampling certain groups), one can directly use the desired distribution to create a custom data order and the model will adapt to it immediately. 

\textit{\textbf{Takeaway 3:} The training data order has an immediate impact on the model's fairness scores. That is, the data distribution in the most recent gradient updates can control the model's group level accuracy in only a single epoch of fine-tuning.}

%% file: sections/sec_applications.tex
\section{Applications of the Impact of Data Order on Fairness}
\label{sec:manipulating}

With a better understanding of the impact of data order on model fairness and how to control it, we now explore some practical applications of our observations.

\subsection{Capturing Fairness Variance in a Single Run}

We now return to our original problem of capturing fairness variance without wasting computing resources on a large number of training runs. We saw an immediate impact of data order on model fairness, which shows that fairness variance across multiple training runs can instead be studied as simply the randomness in data order at their last epochs. As these orders randomly reshuffled, their distribution across multiple trainings should be the same as their distribution across epochs in a single training run. Thus, we propose evaluating fairness of intermediate checkpoints in a single training run as a proxy for multiple runs.

To test the similarity in both distributions, we simply plot the distribution of fairness scores for checkpoints across 200 training runs (for three different stopping epochs) and across epochs $100$ to $300$ in a single training run (for three different training runs) in Fig. \ref{fig:variance_single_run}(a). 
Clearly, the empirical distribution of fairness across multiple training runs closely matches the distribution across a single training run. We also perform the Kolmogorov–Smirnov (KS) test \citep{massey1951kolmogorov} to match sampling across multiple runs (at epoch 300) and sampling across epochs (for a single training run with seed 0). The maximum difference in empirical CDF of the two distributions was only $0.07$, and the KS test gave the p-value of $0.712$, i.e. the probability of the hypothesis that both set of fairness scores were sampled from the same underlying distribution.

\begin{figure*}[hbtp]
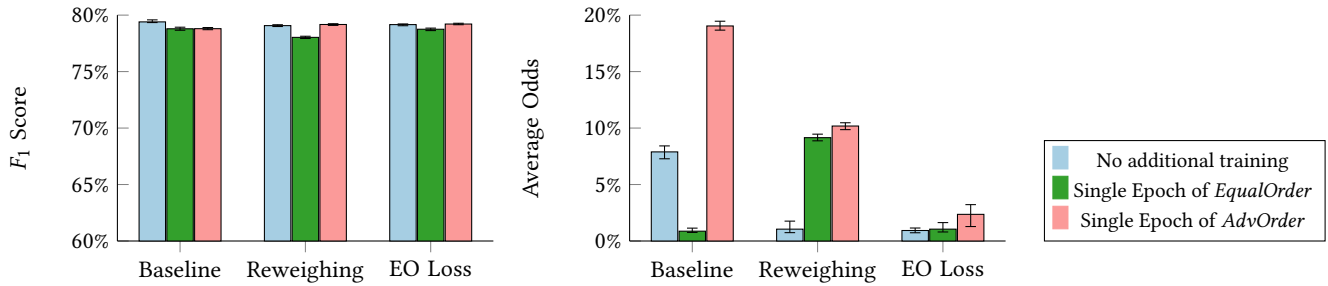

	\centering
	\input{figure_scripts/fig_sota} 
	\input{figure_scripts/fig_sota_legend} 
    \caption{\textbf{Comparing data order manipulation with other bias mitigation algorithms:} Under baseline training setup, by changing the data order for just a single epoch of fine-tuning, \textit{EqualOrder} gets competitive performance to commonly used bias mitigation methods. Similarly, \textit{AdvOrder} gets significantly worse fairness than even the baseline. For Reweighing, we see an increase in bias even with \textit{EqualOrder}, as the beneficial weighing of minority creates bias towards the majority. On the other hand, even reweighing is not enough to counter the effects of \textit{AdvOrder}. Finally, equalized odds loss is capable of dynamically adapting to the model's changing predictive behavior, yet we still observe increase in bias under \textit{AdvOrder}.}
	\label{fig:sota}
\end{figure*}

Furthermore, we also test the quality of black swans, i.e. the best models under certain quality measure, as a function of number of unique training runs and number of epochs evaluated per training run, in Fig. \ref{fig:variance_single_run}(b).
For all $(t, s) \in [1, 50]$, we perform $s$ unique training runs (while changing both forms of randomness, i.e., weight initialization and random reshuffling), and evaluate the model for last $t$ epochs in each training run, thus accumulating a total of $t*s$ checkpoints. We then calculate two different quality measures for these set of checkpoints, i.e., \textit{(i)} the best fairness achieved across all checkpoints, and \textit{(ii)} the Hausdorff distance \citep{birsan2005one} of the Pareto-front (including both fairness and $F_1$ scores) from the best achievable Pareto-front, i.e., for $t=50;s=50$. Finally, the experiment is repeated and averaged $50$ times to compensate for randomness in the $s$ training runs.
Interestingly, the black swans for both quality measures show no significant distinction between increasing the number of training runs or evaluating multiple epochs per training run, i.e., sampling more checkpoints in either direction gives us similar improvements.

It is clear from our results that the commonly used method to capture fairness variance in literature ($t=1;s=50$) is highly inefficient use of computing resources, and one can extract the same quality of black swans (and overall fairness variance) by simply observing fairness across multiple checkpoints in a single training run ($t=50;s=1$), which would require $50$ times less computation.
With these experiments, we showed direct benefits of evaluating multiple epochs in a single training run, saving huge amounts of resources and time in capturing the overall fairness variance.

        \textit{\textbf{Takeaway 4:} Fairness distribution across multiple runs is empirically the same as that across epochs within a single run.}

\subsection{Bias Mitigation via Data Order Manipulation}
\label{subsec:manipulation_mitigation}
To measure the effectiveness of our group accuracy manipulation, we extend the discussion from Section \ref{subsec:manipulating} for two special ratios, $1:1$ and $1:3$ between positive and negative labels of group Female (for ACSIncome dataset), and call them \textit{EqualOrder} and \textit{AdvOrder} respectively. 
More specifically, we fine-tune converged models with a single epoch of \textit{EqualOrder} (and \textit{AdvOrder}), and record $F_1$ score and average odds in Fig. \ref{fig:sota}.
We perform experiments with three unique setups, (i) Baseline training, (ii) Reweighing \citep{kamiran2012data}, a data pre-processing which weighs every label-group pair based on its representation in the overall dataset, and (iii) Equalized Odds Loss \citep{fukuchi2020fairtorch}, an in-processing loss function to nudge the model towards fair predictions.
By training with \textit{EqualOrder} for a single epoch, the baseline model achieves competitive fairness scores to other bias mitigation algorithms. 
On the other hand, using \textit{AdvOrder} can further push the model bias, emphasizing the adversarial power of data ordering, even in presence of explicit mitigation algorithms. 

Notably, reweighing suffers from an unexpected high bias even under \textit{EqualOrder}, as the combination of ideally distributed data order suffix along with increase in the minority data weights pushes the model towards significant unfair behavior against the majority (as opposed to against the minority in all other results). Moreover, equalized odds loss shows controlled damage under \textit{AdvOrder} due to the loss function regularly adapting to the degrading behavior, but the unfairness still increases. \textit{AdvOrder} is dangerous as it still maintains the overall accuracy, but favors the majority.
We can force even worse fairness gaps by pushing the ratio to its extreme, however that will impact the model's overall accuracy. 
These results cement the effectiveness of manipulating group level accuracy by controlling the data order for just a single epoch of fine-tuning.

        \textit{
        \textbf{Takeaway 5:} A data order with a balanced suffix can significantly improve in fairness scores. Similarly, even bias mitigation algorithms can fail when trained with an adversarial data order.
        }

%% file: sections/sec_conclusion.tex
\section{Conclusion}
\label{sec:conclusion}

Fairness variance due to changing randomness in deep learning has raised concerns regarding the reliability of existing results in literature \cite{amir2021impact,sellam2021multiberts,soares2022your}. In our work, we took a close look at various sources of randomness, and found a dominant impact of data order on model fairness, which we showed was in turn due to a higher prediction uncertainty of the trained model on under-represented groups in the dataset. We further demonstrated that the distribution seen by the model in the most recent gradient updates can be easily exploited to achieve desirable group-level accuracy behavior, and proposed several practical applications of this immediate impact of data order on model fairness, including a highly efficient alternative to executing multiple training runs when studying fairness variance due to randomness in training.

In our work, we focused only on the discrete decisions made by the model, as we were investigating the impact of non-determinism in model training on its fairness. However, further extensions of this discussion to trends in the internal state of the learned model can reveal even granular characteristics,
and has potential application in similar fields of research, for example, understanding high variance in out-of-distribution generalization \cite{mccoy2020berts}, exploiting model multiplicity under various settings \cite{black2022model}, and many more.


%% file: sections/sec_appendix_intro.tex
\section{Non-Determinism in Model Training}
\label{sec:app_non_determinism}

In our paper, we focus on fairness variance due to randomness in the training algorithm (i.e., weight initialization and random reshuffling). To control the randomness, we set manual seeds at various intermediate locations in our code. We refer to the seed set right before building the neural network as the weight initialization seed, which influences the randomness in sampling the weight values. Similarly, we refer to the seed set right before the first training data shuffling as random reshuffling seed, which influences the data order that will be used as reference for the rest of the training. During training, we simply set the epoch number as seed right before reshuffling the reference data order at every epoch. As we can change the reference data order by changing the random reshuffling seed, this setup allows us to control non-determinism in data order throughout training with a single random seed.

We also perform experiments with dropout regularization during training.
The epoch number seed set at the start of every epoch serves dual functionality, and also controls the randomness from dropout regularization. That is, fixing both weight initialization and random reshuffling seeds allows us to deterministically replicate model training, while changing both seeds simultaneously is similar to the discussion of non-determinism currently present in fair deep learning literature. 

% ...

\section{Datasets}
\label{sec:app_datasets}

\paragraph{ACSIncome}
ACSIncome is one of the five pre-defined tasks in the Folktables dataset \cite{ding2021retiring}, which was recently collected to improve the older and commonly used UCI Adult Income dataset \cite{Dua:2019}. More specifically, we use the subset of Folktables dataset from the state of California, USA for 2018. The dataset contains a total of $195,665$ data points with $10$ features each, where each data points represents an individual. The task is a binary classification to predict whether the individual's income is above $\$50,000$. For fairness measures, we use perceived gender (\textit{Sex}) as the sensitive attribute, which is also one of the $10$ input features.

List of features : \textit{Age}, \textit{Class of worker}, \textit{Educational attainment}, \textit{Marital status}, \textit{Occupation}, \textit{Place of birth}, \textit{Relationship}, \textit{Usual hours worked per week in past 12 months}, \textit{Sex}, \textit{Recoded detailed race code}

\paragraph{ACSEmployment} ACSEmployment is another one of the five pre-defined tasks in the Folktables dataset \cite{ding2021retiring}. We use the same subset from the state of California, USA for 2018 as above. The dataset contains a total of $378,817$ data points with $16$ features each, and the task is a binary classification to predict whether the individual is employed or not. For fairness measures, same as above, we use perceived gender (\textit{Sex}) as the sensitive attribute.

List of features : \textit{Age}, \textit{Educational attainment}, \textit{Marital status}, \textit{Sex}, \textit{Disability recode}, \textit{Employment status of parents}, \textit{Mobility status}, \textit{Citizenship status}, \textit{Military service}, \textit{Ancestry recode}, \textit{Nativity}, \textit{Relationship}, \textit{Hearing difficulty}, \textit{Vision difficulty}, \textit{Cognitive difficulty}, \textit{Recoded detailed race code}, \textit{Grandparents living with grandchildren}

\paragraph{CelebA} CelebA dataset is a large scale celebrity face attributes dataset \cite{liu2015faceattributes}, which contains a total of $202,599$ images of celebrities with $40$ different binary labels each. We focus on the binary classification task of the 'smiling' label in the dataset, while we use the 'gender' label as the sensitive attribute for fairness evaluation. Moreover, we do not directly use the images of CelebA dataset, but instead pass them through a pre-trained and frozen ResNet-50 backbone \cite{he2016deep} to extract image representations, which are treated as inputs to our model.

\section{Training Curve and Convergence}
\label{sec:app_training_curve}

\begin{figure*}[htbp]
	\centering
    \input{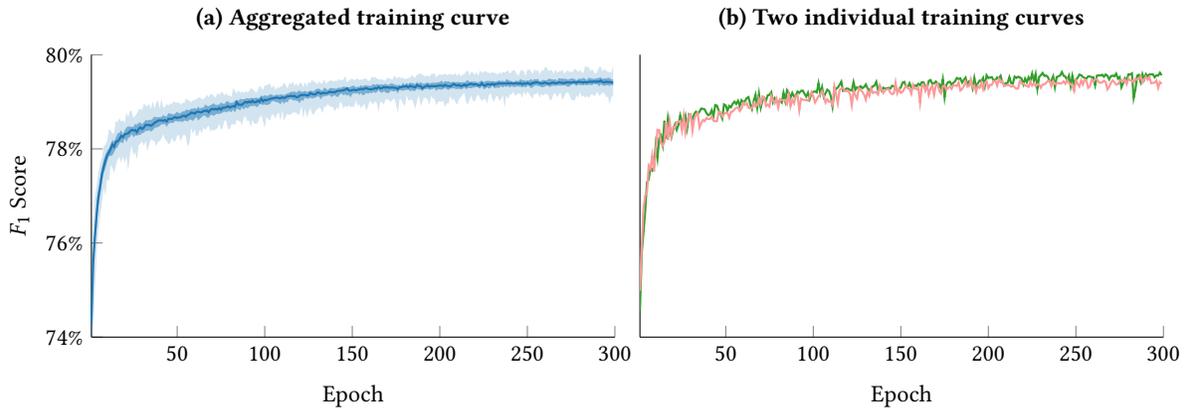}
    \caption{(Left) Aggregated training curve for all 50 training runs with both changing weight initialization and random reshuffling. (Right) Example of two individual training runs.}
	\label{fig:training_curve}
\end{figure*}

To understand why we choose to study the model behavior between epochs $100$ and $300$, we point the reader towards the overall training curve of the model plotted in Fig. \ref{fig:training_curve}. It is clear that the model has converged by epoch $100$, and maintains stable accuracy scores for the last $200$ epochs.

\begin{figure*}[htbp]
	\centering
	\scalebox{0.9}{\input{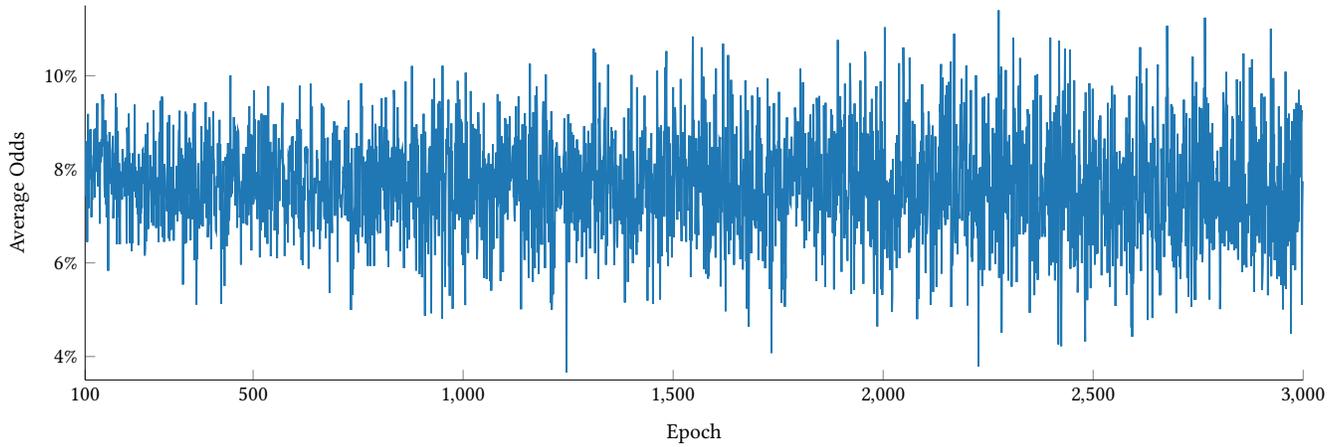}}
    \caption{A single training run extended to $3000$ epochs.}
	\label{fig:training_curve_3000}
\end{figure*}

Even though the accuracy scores have converged, we still find high variance in fairness scores as discussed in the main text of the paper. One might suspect this implies that fairness scores could take longer to converge. To check this, we allow a single training to run for a total of $3000$ epochs (as opposed to the standard $300$ epochs used in all other experiments in our paper) and collect the fairness scores in Fig. \ref{fig:training_curve_3000}. It is clear that the model fairness does not stabilize by simply increasing the number of epochs, which further supports our hypothesis of never-ending local oscillations due to SGD noise that cause high fairness variance.

\section{Random Order for Immediate Impact}
\label{sec:app_random_order}

\begin{figure*}[hbtp]
	\centering
    \input{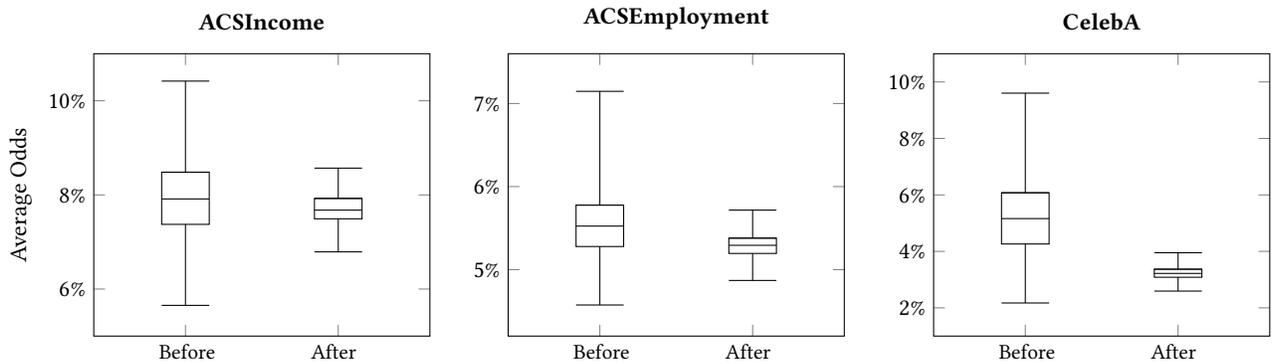}
    \caption{(\textit{Before}) Multiple checkpoints taken from training runs with changing weight initialization, training data order, and even number of training epochs, show a high range of fairness variance (i.e., average odds variance), as expected. \textit{(After)} However, these fairness scores are stabilized significantly by training these checkpoints for only a single epoch on a common randomly chosen data order. This shows the immediate impact of data order on model fairness.}
	\label{fig:immediate_data_order}
\end{figure*}

We used carefully chosen data order to show the immediate impact of data order in Section \ref{sec:data_order}. Here, we provide additional results on randomly chosen data order to show that the property does hold even for a random data order.
Sae=me as in the original experiment, we sample 1000 unique checkpoints randomly from last 200 epochs of 50 different training runs while allowing both forms of training non-determinism simultaneously. We then fine-tune each of these checkpoints for exactly one epoch on a common, ad this time randomly chosen, data order. We collect fairness variance across checkpoints before and after this single epoch of training in Fig. \ref{fig:immediate_data_order}.

Before training on the same data order for a single epoch, these checkpoints represent the complete range of fairness variance previously noted. However, after only a single epoch of training on a common data order, these models have all moved towards the same fairness score with significantly lower variance. This highlights the immediate impact of data order on model fairness, which is stable based on only the data order of the most recent epoch.

\section{Sanity Check for Data Order Suffix}
\begin{figure*}[htbp]
    \centering
    \input{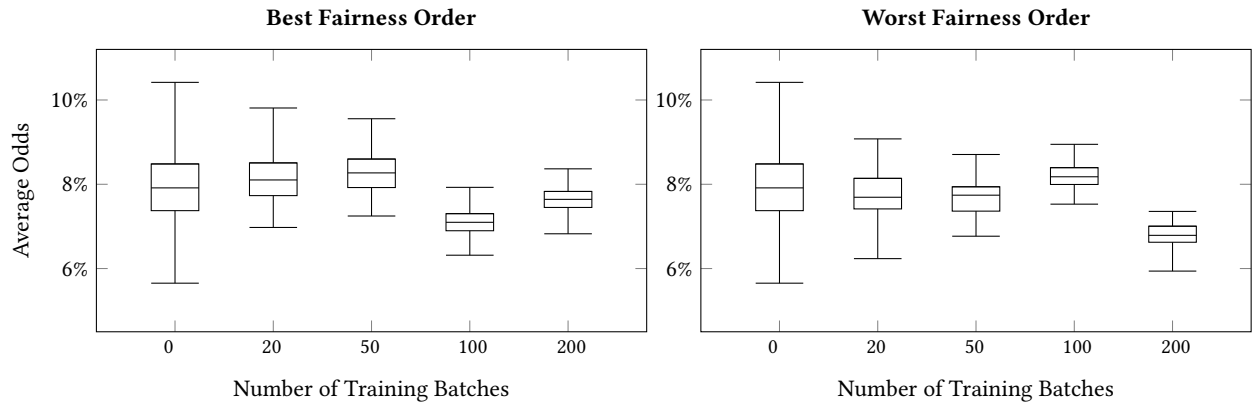}
    \caption{Additional experiments by choosing $b$ batches, similar to the experiments in Figure \ref{fig:suffix_shuffle}, but choosing batches randomly instead of the suffix. The results still maintain stability as $b$ increases but now they stabilize to random fairness scores instead of best and worst fairness scores as seen in Figure \ref{fig:suffix_shuffle}. Clearly, its the most recent batches in that order which truly governs the model fairness.}
    \label{fig:app_random_shuffle}
\end{figure*}

We showed that the fairness scores are dominated by the most recent gradient updates as seen by the model. As a sanity check, we also provide ablation study for experiments in Fig. \ref{fig:suffix_shuffle}, but choose the $b$ batches randomly instead of from the suffix. The results are collected in Fig. \ref{fig:app_random_shuffle}. The results show while the fairness variance does get smaller with more common batches (as expected, see also Fig. \ref{fig:immediate_data_order}), its predictability is lost when choosing the $b$ batches randomly. Thus, it is indeed the batches from data order suffix that govern the predictability of model fairness.

%% file: sections/sec_appendix_datasets.tex
\section{Additional Experiments for All Datasets}
\label{sec:app_all_datasets}

\subsection{Weight Initialization and Random Reshuffling}\label{app:wghtvsshuffle}

\begin{figure*}[htbp]
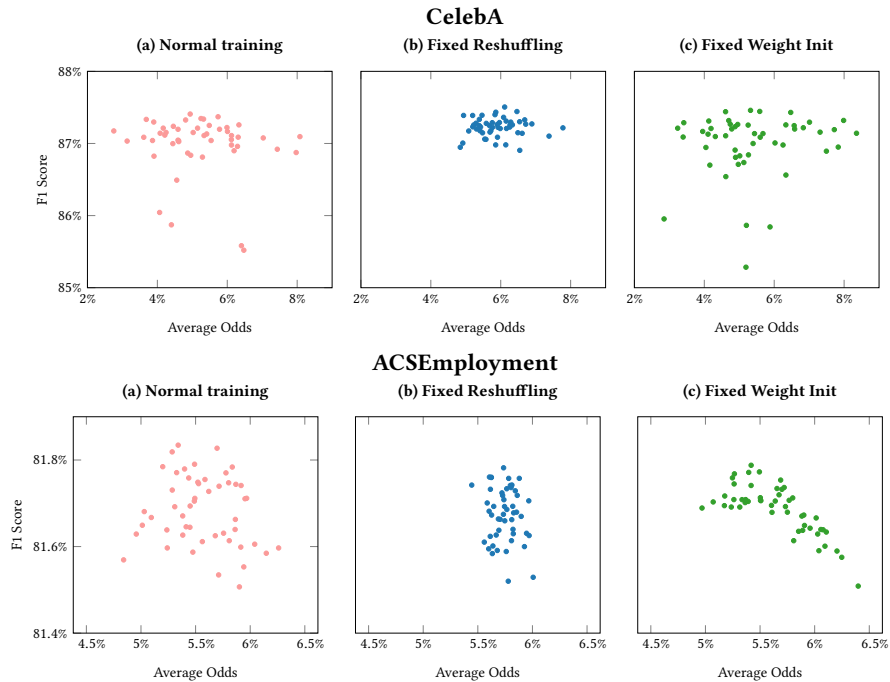

	\centering
	\begin{tabular}{c}
	\textbf{CelebA} \\
    \scalebox{0.7}{\input{figure_scripts/appendix/fig_decouple_scatter_celeba}} \\
    \textbf{ACSEmployment} \\
    \scalebox{0.7}{\input{figure_scripts/appendix/fig_decouple_scatter_acsemployment}} \\
	\end{tabular}
    \caption{Additional experiments on CelebA and ACSEmployment datasets reveal similar trends as seen in Figure \ref{fig:decouple_scatter}. Random reshuffling of data order is the dominant source of variance in fairness scores, with very little influence from the weight initialization.}
	\label{fig:app_decouple_scatter}
\end{figure*}

\begin{figure*}[htbp]
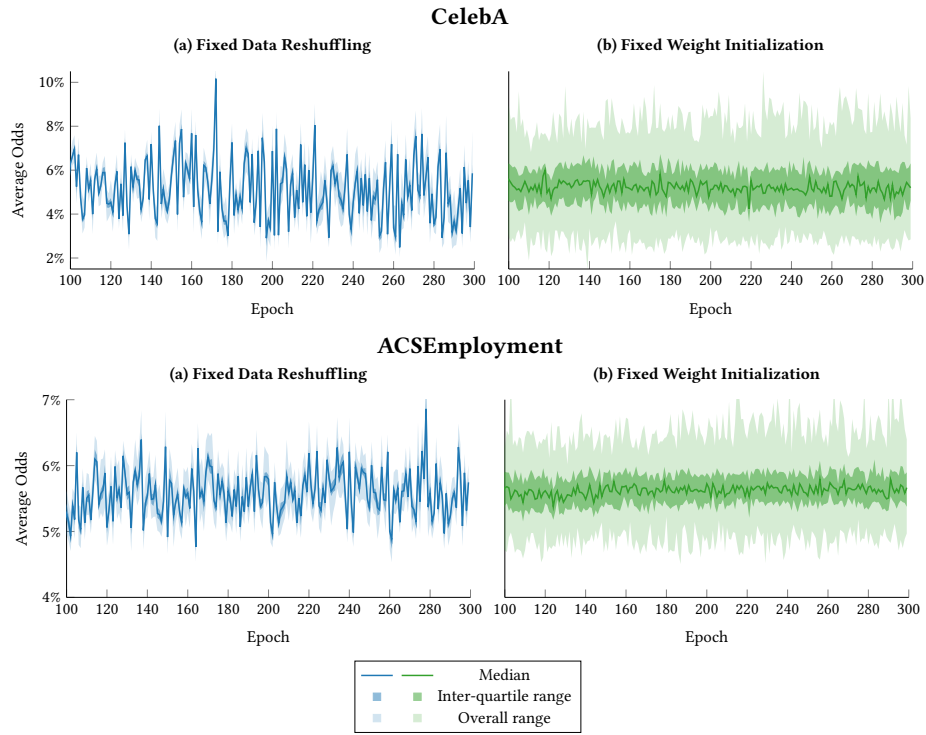

	\centering
	\begin{tabular}{c}
	\textbf{CelebA} \\
	\scalebox{0.7}{\input{figure_scripts/appendix/fig_curve_correlation_celeba}} \\
	\textbf{ACSEmployment} \\
	\scalebox{0.7}{\input{figure_scripts/appendix/fig_curve_correlation_acsemployment}} \\
	\scalebox{0.7}{\input{figure_scripts/fig_curve_correlation_legend}} \\
	\end{tabular}
    \caption{Additional experiments on CelebA and ACSEmployment datasets reveal similar trends as in Figure \ref{fig:decouple_correlation}. These results further highlight the dominant impact of random reshuffling on fairness.}
	\label{fig:app_decouple_correlation}
\end{figure*}

We provide additional results in Fig. \ref{fig:app_decouple_scatter}, \ref{fig:app_decouple_correlation} on CelebA and ACSEmployment to show the dominance of reshuffling.

\newpage
\subsection{Changing Predictions and Data Distribution}
\label{subsec:app_data_distrib}

\begin{figure*}[hbtp]
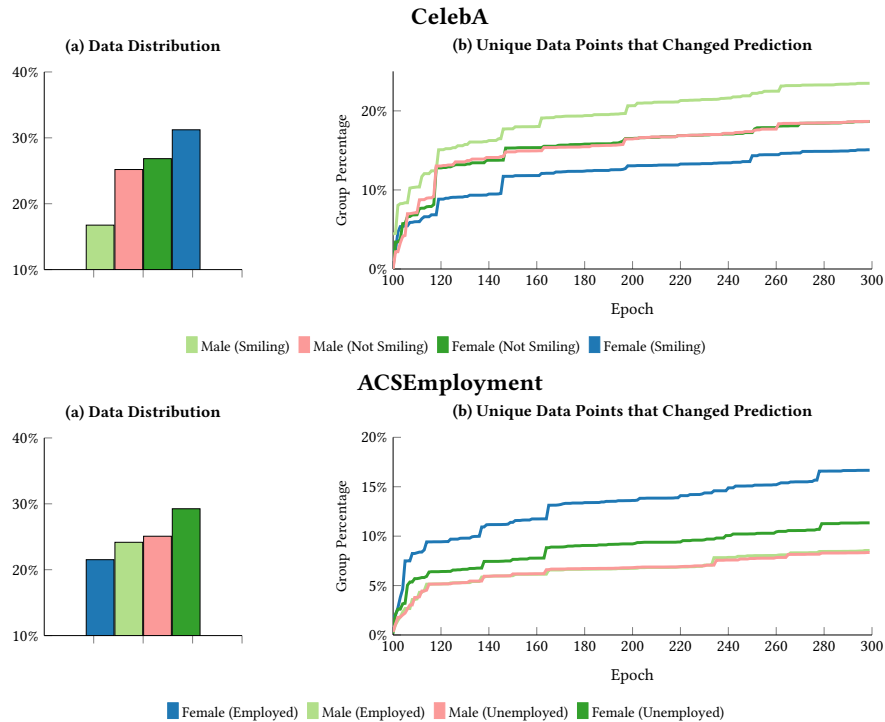

	\centering
	\begin{tabular}{cc}
	\multicolumn{2}{c}{\textbf{CelebA}} \\
    \scalebox{0.7}{\input{figure_scripts/appendix/fig_data_distrib_celeba}} &
    \scalebox{0.7}{\input{figure_scripts/appendix/fig_example_forgetting_total_celeba}} \\
    \multicolumn{2}{c}{\scalebox{0.7}{\input{figure_scripts/appendix/fig_example_forgetting_legend_celeba}}} \\
	\multicolumn{2}{c}{\textbf{ACSEmployment}} \\
    \scalebox{0.7}{\input{figure_scripts/appendix/fig_data_distrib_acsemployment}} &
    \scalebox{0.7}{\input{figure_scripts/appendix/fig_example_forgetting_total_acsemployment}} \\
    \multicolumn{2}{c}{\scalebox{0.7}{\input{figure_scripts/appendix/fig_example_forgetting_legend_acsemployment}}} \\
    \end{tabular}
    \caption{Additional experiments on CelebA and ACSEmployment datasets reveal similar trends as seen in Figure \ref{fig:example_forgetting}. These results highlight subgroups with least representation being the most vulnerable to changing predictions.}
	\label{fig:app_example_forgetting}
\end{figure*}

We show the relationship between data distribution and changing predictions for CelebA and ACSEmployment in Fig. \ref{fig:app_example_forgetting}. The group with least representation maintains being the most vulnerable to changing predictions.

\subsection{Capturing Variance in a Single Training Run}

\begin{figure*}[hbtp]
    \centering
    \begin{minipage}{\textwidth}
        \centering
        \textbf{CelebA} \\
    \end{minipage} \\[0.2cm]
    \begin{minipage}{0.65\textwidth}
        \centering
        \textbf{(a) Fairness Score Distribution}\\[0.2cm]
        \includegraphics[width=\linewidth]{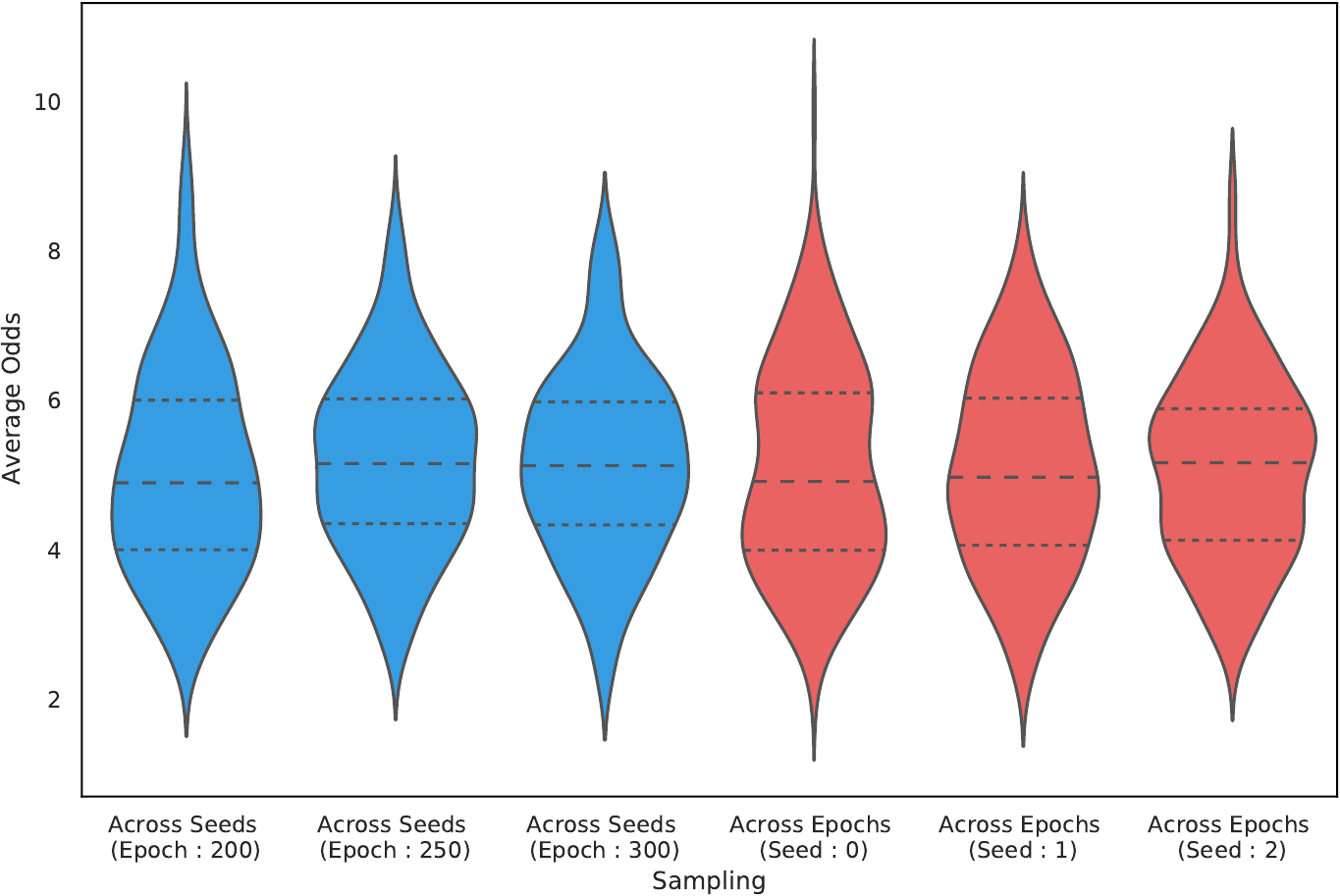}
    \end{minipage}\hfill
    \begin{minipage}{0.28\textwidth}
        \centering
        \textbf{(b) Black Swans} \\[0.2cm]
        \includegraphics[width=\linewidth]{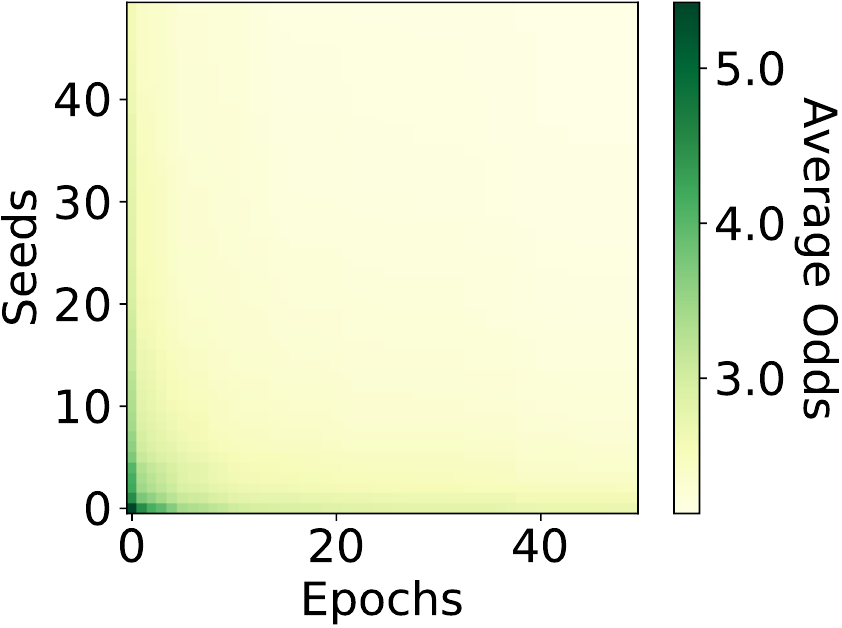} \\[0.2cm]

        \includegraphics[width=\linewidth]{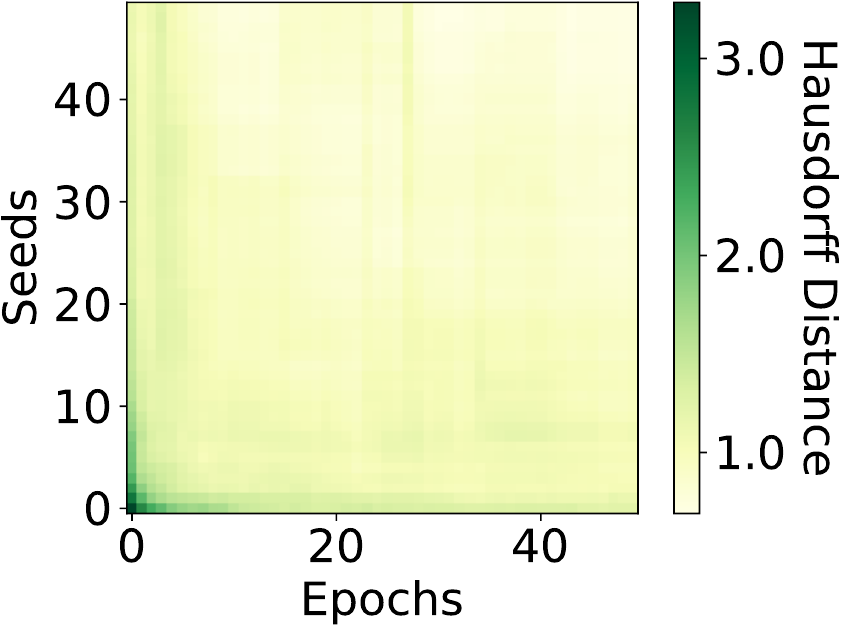}
    \end{minipage}\hfill
    \vspace{2em}
    \begin{minipage}{\textwidth}
        \centering
        \textbf{ACSEmployment} \\
    \end{minipage} \\[0.2cm]
    \begin{minipage}{0.65\textwidth}
        \centering
        \textbf{(a) Fairness Score Distribution}\\[0.2cm]
        \includegraphics[width=\linewidth]{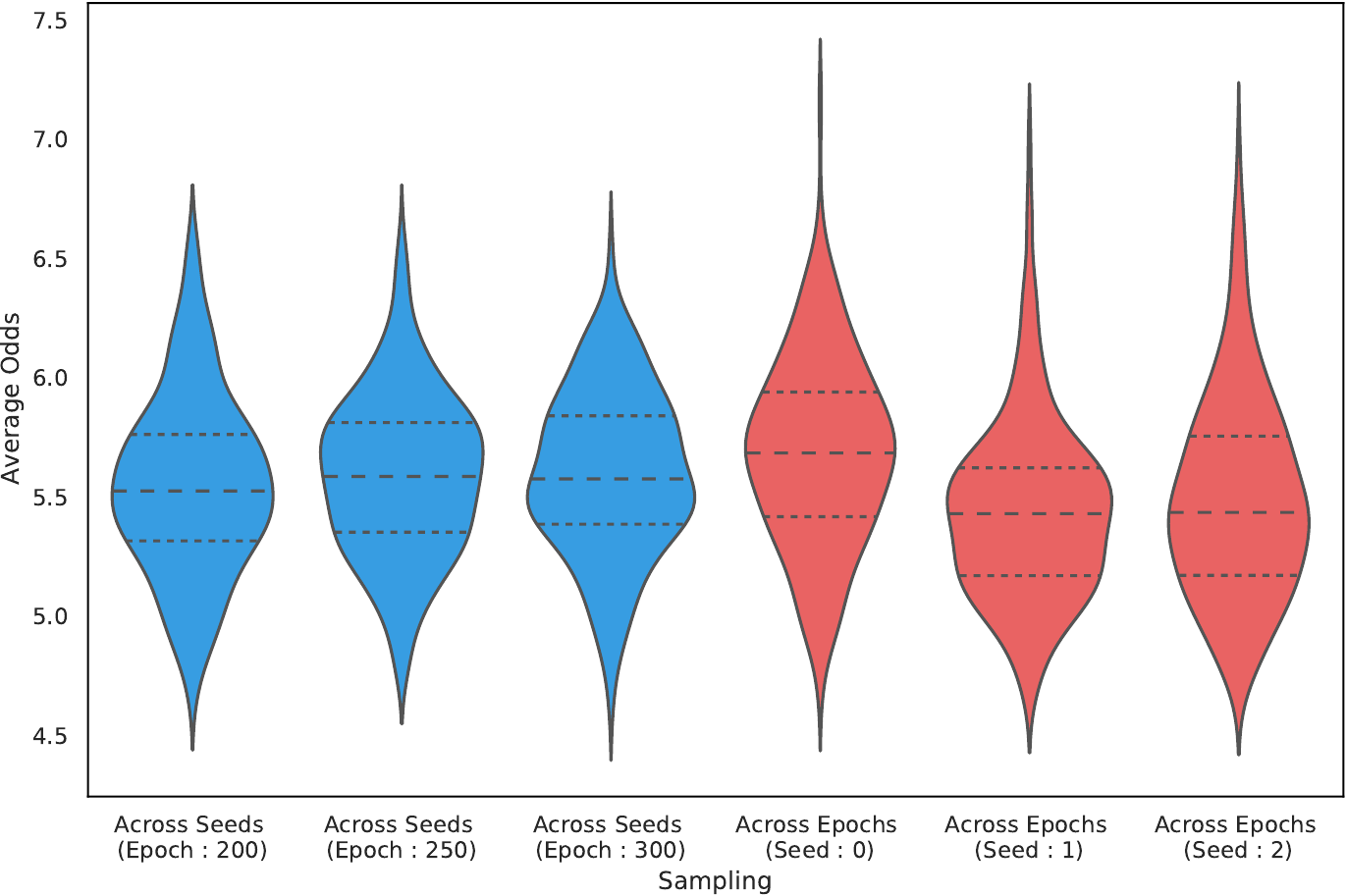}
    \end{minipage}\hfill
    \begin{minipage}{0.28\textwidth}
        \centering
        \textbf{(b) Black Swans} \\[0.2cm]
        \includegraphics[width=\linewidth]{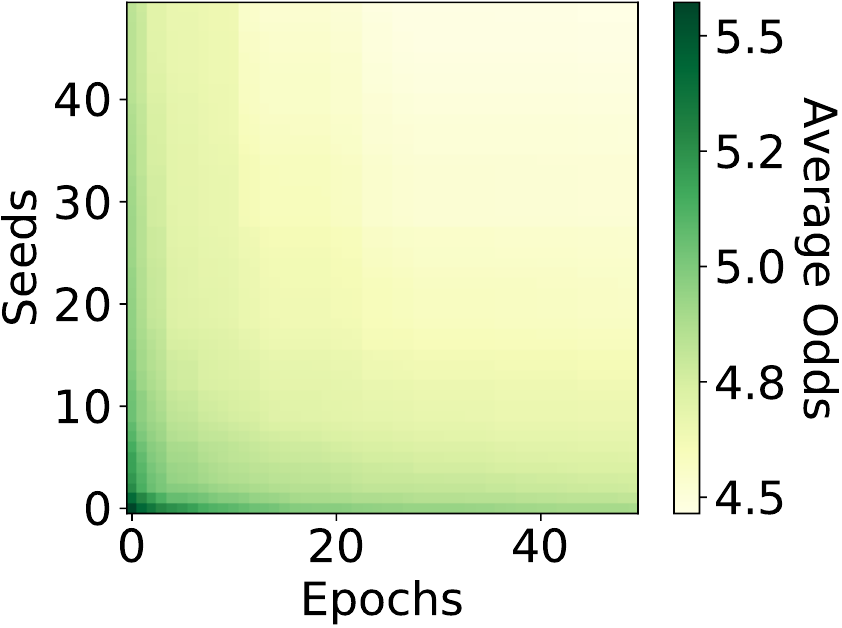} \\[0.2cm]

        \includegraphics[width=\linewidth]{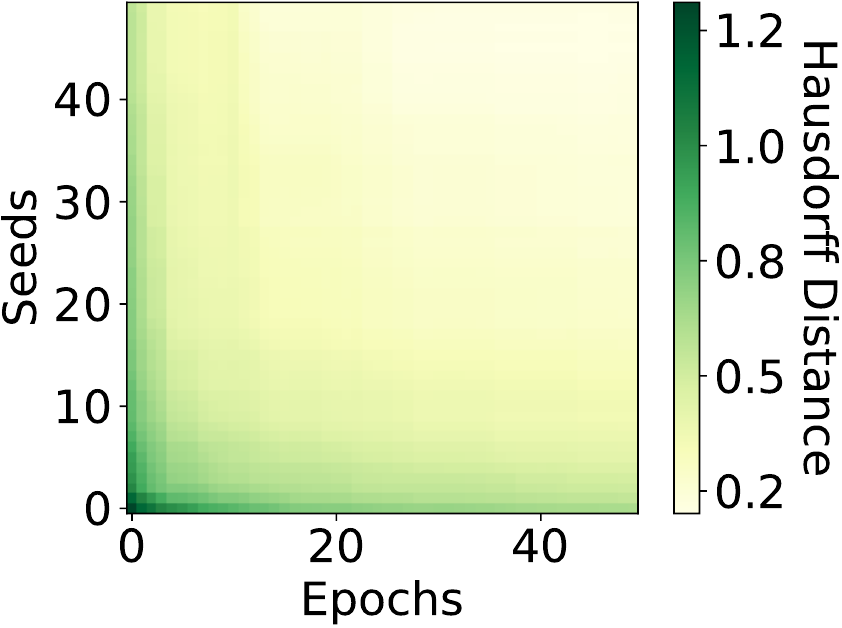}
    \end{minipage}\hfill \\[0.3cm]
    \caption{Additional experiments on CelebA and ACSEmployment datasets reveal similar trends as seen in Figure \ref{fig:variance_single_run}. Fairness scores (average odds) across multiple training runs and across epochs in a single training run have similar empirical distributions. Thus, studying this distribution across epochs provides a highly efficient alternative to executing multiple training runs.}
    \label{fig:app_variance_single_run}
\end{figure*}

We show the empirical similarity of distribution across multiple training runs and multiple epochs in a single training run for additional datasets CelebA and ACSEmployment in Fig. \ref{fig:app_variance_single_run}.

\newpage
\subsection{Manipulating Group Level Accuracy with Data Order}

\begin{figure*}[htbp]
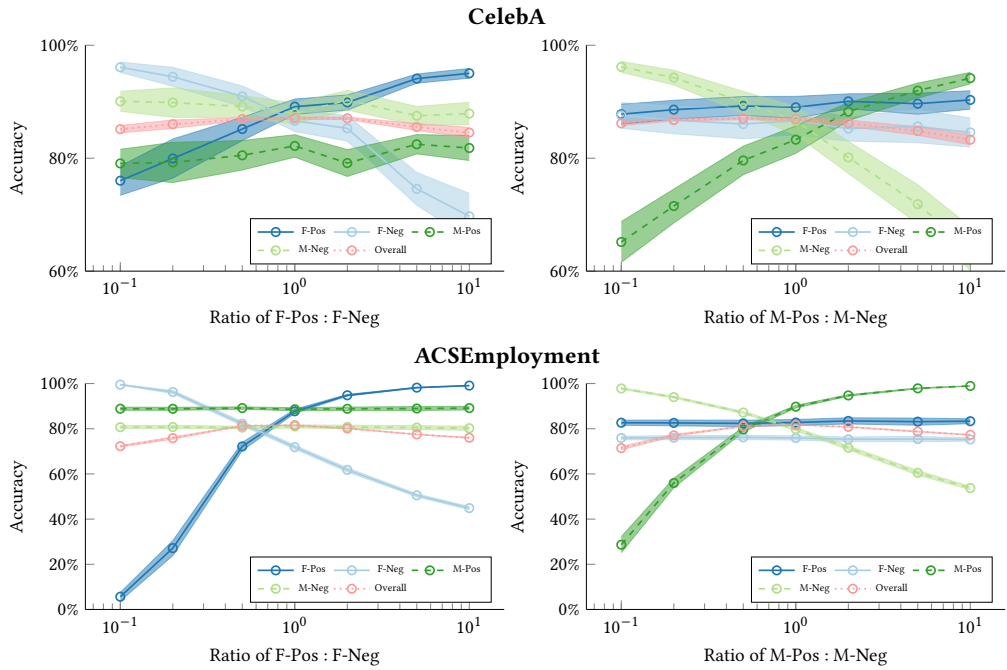

	\centering
    \begin{tabular}{c}
    \textbf{CelebA} \\
    \scalebox{0.8}{\input{figure_scripts/appendix/fig_manipulate_celeba}} \\
    \textbf{ACSEmployment} \\
    \scalebox{0.8}{\input{figure_scripts/appendix/fig_manipulate_acsemployment}} \\
    \end{tabular}
    \caption{Additional experiments on CelebA and ACSEmployment datasets reveal similar trends as seen in Figure \ref{fig:manipulate}. In only a single epoch of training, we are able to manipulate the group level accuracy trade-off, with relatively small impact on overall accuracy.
    }
	\label{fig:app_manipulate}
\end{figure*}

\begin{figure*}[hbtp]
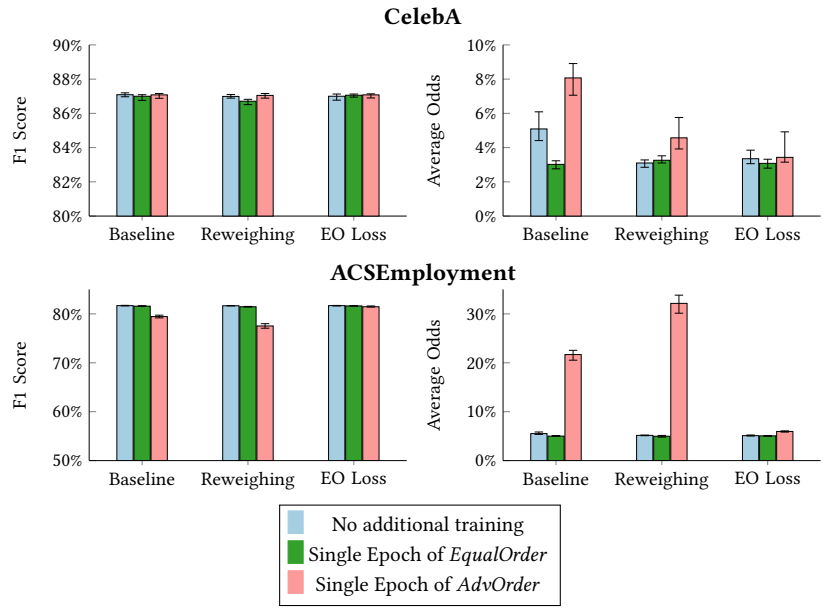

	\centering
	\begin{tabular}{c}
	\textbf{CelebA} \\
	\scalebox{0.8}{\input{figure_scripts/appendix/fig_sota_celeba}} \\
	\textbf{ACSEmployment} \\
	\scalebox{0.8}{\input{figure_scripts/appendix/fig_sota_acsemployment}} \\
	\input{figure_scripts/fig_sota_legend} \\
	\end{tabular}
    \caption{Additional experiments on CelebA and ACSEmployment reveal similar trends as seen in Figure \ref{fig:sota}. \textit{EqualOrder} gets competitive performance to commonly used bias mitigation methods. Similarly, \textit{AdvOrder} gets significantly worse fairness than even the baseline.}
	\label{fig:app_sota}
\end{figure*}

We provide additional results in Figure \ref{fig:app_manipulate}, \ref{fig:app_sota}, highlighting the predictability of model fairness based on the data order.

%% file: sections/sec_appendix_hyperparameters.tex
\section{Additional Experiments for Changing Hyperparameters}
\label{sec:app_hyperparameters}

We provide additional experiments for \textit{(i)} batch size $16$ and $1024$, deviating from the default batch size of $128$ in the main text, \textit{(ii)} learning rate $0.01$ and $0.0001$, deviating from the default learning rate of $0.001$ in the main text, and \textit{(iii)} model architecture with two hidden layers containing $2048$ and $64$ neurons respectively, deviating from the single hidden layer architecture used in the main text.

\subsection{Training Curves and Convergence}
\begin{figure*}[htbp]
	\centering
	\scalebox{0.7}{\input{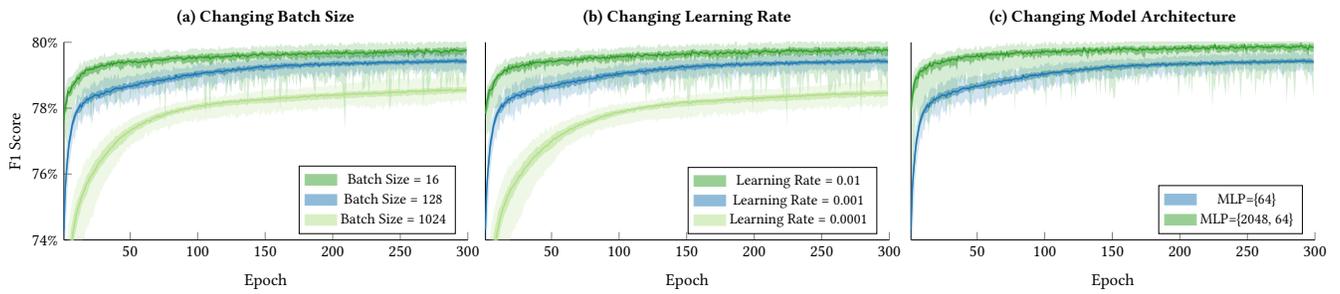}}
    \caption{Training curve under changing batch size, learning rate, and model architecture. It is clear that decreasing the batch size, increasing the learning rate, or using a bigger model architecture results in a faster model convergence but with higher variance even in accuracy scores. On the other hand, models with high batch size or low learning rates tend to not achieve the same accuracy scores and have not reached convergence even at $300$ epochs.}
	\label{fig:training_curve_all}
\end{figure*}

We first observe the training curve in all the settings described above, along with the original setting, in Fig. \ref{fig:training_curve_all}. It is clear that models with a bigger batch size (or a smaller learning rate) do not converge to the same F1 scores as other models. This further indicates the need of randomness and noise in the learning algorithm to give neural networks more 'exploration energy' and facilitate better and faster convergence. On the other hand, a smaller batch size (or a larger learning rate) comes with excessive variance in the model's F1 score, which highlights the importance of achieving an appropriate balance between several hyperparameter choices.

\begin{figure*}[htbp]
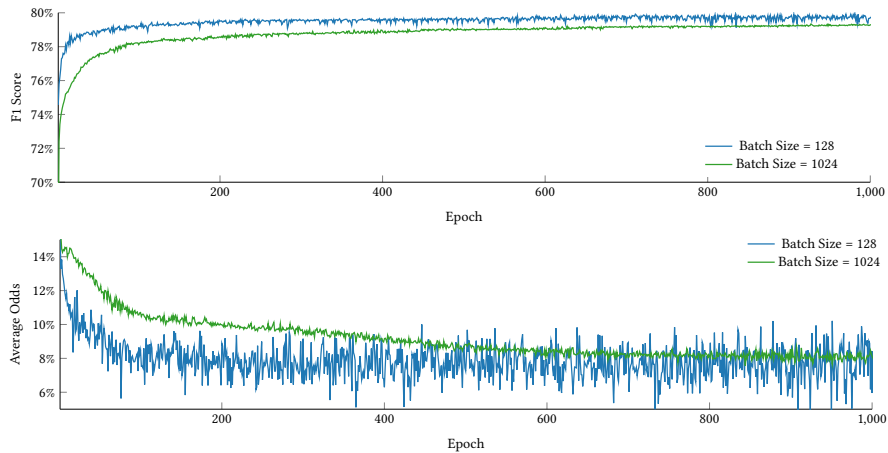

	\centering
	\scalebox{0.6}{\input{figure_scripts/appendix/fig_training_curve_bs_1024_fscore}}
	\scalebox{0.6}{\input{figure_scripts/appendix/fig_training_curve_bs_1024_avgodds}}
    \caption{Using a higher batch size can over time achieve stable fairness scores, however the convergence speed is significantly slower. Moreover, it looses a noticeable margin of F1 score.}
	\label{fig:app_training_curve_bs_1024}
\end{figure*}

\begin{figure*}[htbp]
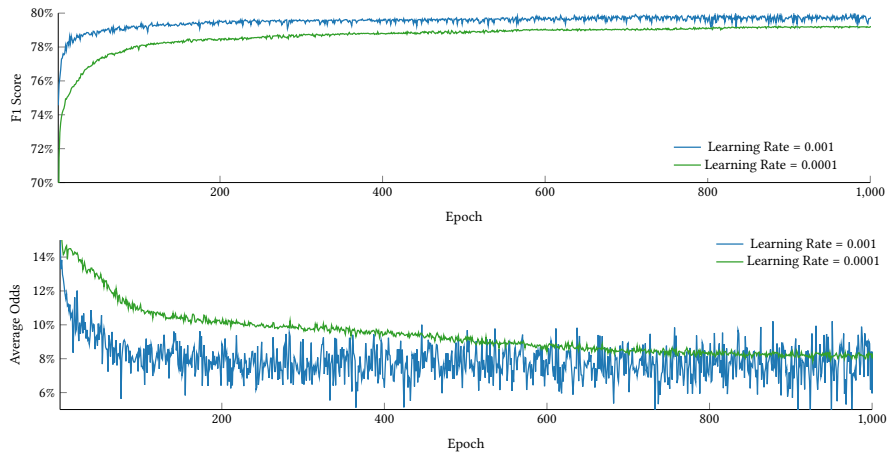

	\centering
	\scalebox{0.6}{\input{figure_scripts/appendix/fig_training_curve_lr_0001_fscore}}
	\scalebox{0.6}{\input{figure_scripts/appendix/fig_training_curve_lr_0001_avgodds}}
    \caption{Using a lower learning rate can over time achieve stable fairness scores, however the convergence speed is significantly slower. Moreover, it looses a noticeable margin of F1 score.}
	\label{fig:app_training_curve_lr_0001}
\end{figure*}

To better understand the difficulty of reaching convergence, we continue training a single instance with batch size $1024$ (and separately with learning rate $0.0001$) for a total of $1000$ epochs and compare it against the standard training setup (i.e., with batch size $128$ and learning rate $0.001$). The results are collected in Fig. \ref{fig:app_training_curve_bs_1024}, \ref{fig:app_training_curve_lr_0001}. It is clear that a larger batch size (or lower learning rate) slows down the convergence speed. Moreover, it also does not achieve the same F1 scores previously seen, possibly due to not being able to take complete advantage of the noisy nature of mini-batch gradient descent. Thus, training a model with hyperparameters that enforces stability (i.e. large batch size or lower learning rate) is an inefficient solution to solving the problem of fairness variance.

\newpage
\subsection{Weight Initialization and Random Reshuffling}

\begin{figure*}[htbp]
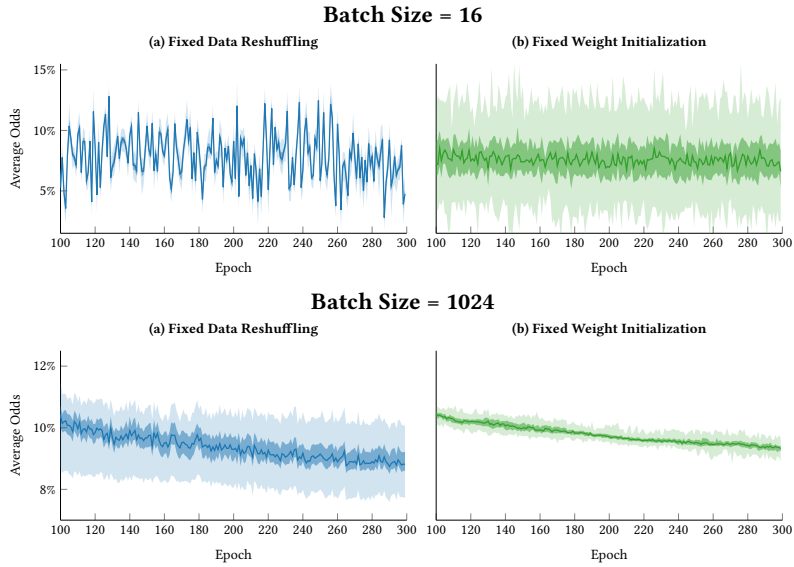

	\centering
	\begin{tabular}{c}
	\textbf{Batch Size = 16} \\
	\scalebox{0.6}{\input{figure_scripts/appendix/fig_curve_correlation_bs_16}} \\
	\textbf{Batch Size = 1024} \\
	\scalebox{0.6}{\input{figure_scripts/appendix/fig_curve_correlation_bs_1024}} \\
	\end{tabular}
    \caption{Additional experiments for changing batch size with experiment setting as in Figure \ref{fig:decouple_correlation}. These results further highlight the dominant impact of random reshuffling on fairness.}
	\label{fig:app_decouple_correlation_hyper_batch}
\end{figure*}

\begin{figure*}[htbp]
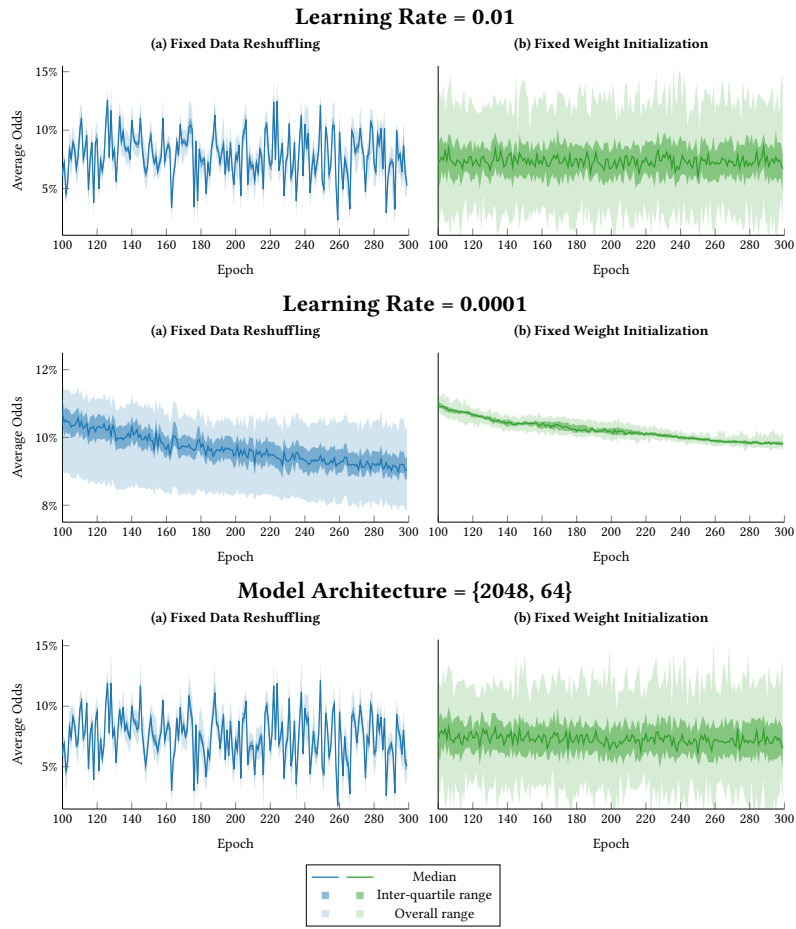

	\centering
	\begin{tabular}{c}
	\textbf{Learning Rate = 0.01} \\
	\scalebox{0.6}{\input{figure_scripts/appendix/fig_curve_correlation_lr_01}} \\
	\textbf{Learning Rate = 0.0001} \\
	\scalebox{0.6}{\input{figure_scripts/appendix/fig_curve_correlation_lr_0001}} \\
	\textbf{Model Architecture = \{2048, 64\}} \\
	\scalebox{0.6}{\input{figure_scripts/appendix/fig_curve_correlation_arch_2048_64}} \\
	\scalebox{0.6}{\input{figure_scripts/fig_curve_correlation_legend}} \\
	\end{tabular}
    \caption{Additional experiments for changing learning rate and model architecture with experiment setting as in Figure \ref{fig:decouple_correlation}. These results further highlight the dominant impact of random reshuffling on fairness.}
	\label{fig:app_decouple_correlation_hyper_lr}
\end{figure*}

We provide results in Fig. \ref{fig:app_decouple_correlation_hyper_batch}, \ref{fig:app_decouple_correlation_hyper_lr} for changing batch size, learning rate, and architecture on ACSIncome dataset to show the dominance of random reshuffling on model fairness.
With a decrease in batch size, the range of fairness variance increases significantly, but the overall expected trends follow the same behavior as noted in the main text, i.e. a sharp change in fairness scores across epochs under fixed data reshuffling, and high variance even for a single epoch across multiple runs under changing data reshuffling (fixed weight initialization). Similar trends can be observed when we increase the learning rate, or provide the training algorithm with a bigger neural model. The increase in variance suggests high instability with smaller batch size, higher learning rate, and bigger neural models, all of which is expected.

On the other hand, one would expect more stable model behavior with a bigger batch size or a smaller learning rate. While this is indeed the case, the comparison here is not fair because as noted earlier (Fig. \ref{fig:training_curve_all}, these models have not yet converged, also evident by the clear downward trend of fairness scores.
It is clear that certain hyperparameter settings are not conducive to efficient convergence, even though they might eventually provide more stable fairness scores.

\newpage
\subsection{Changing Predictions and Data Distribution}
\label{subsec:app_data_distrib}

We note changing predictions for all 5 settings described above in Fig. \ref{fig:app_example_forgetting_hyper}. Note that since we are focusing on changing training hyperparameters for ACSIncome, the data distribution remains the same as in Fig. \ref{fig:example_forgetting}(a) (also copied here in Fig. \ref{fig:app_example_forgetting_hyper} for reference).

Same as before, model instability has increased with smaller batch size, higher learning rate, or bigger model architecture, but the trends of vulnerability for various subgroups remains the same. Even though we see same trends of higher vulnerability for a higher batch size or smaller learning rate, we know that these models have not converged and thus would recommend not making any inference from these two set of results.

\begin{figure*}[hbtp]
	\centering
	\begin{tabular}{cc}
	\multicolumn{2}{c}{\textbf{Batch Size}} \\
	Batch Size = 16 & Batch Size = 1024 \\
    \scalebox{0.7}{\input{figure_scripts/appendix/fig_example_forgetting_total_bs_16}} &
    \scalebox{0.7}{\input{figure_scripts/appendix/fig_example_forgetting_total_bs_1024}} \\
	\multicolumn{2}{c}{\textbf{Learning Rate}} \\
	Learning Rate = 0.01 & Learning Rate = 0.0001 \\
    \scalebox{0.7}{\input{figure_scripts/appendix/fig_example_forgetting_total_lr_01}} &
    \scalebox{0.7}{\input{figure_scripts/appendix/fig_example_forgetting_total_lr_0001}} \\
    \textbf{ACSIncome Data Distribution} & \textbf{Model Architecture = \{2048, 64\}} \\
    \scalebox{0.7}{\input{figure_scripts/fig_data_distrib}} & \scalebox{0.7}{\input{figure_scripts/appendix/fig_example_forgetting_total_arch_2048_64}} \\
    \multicolumn{2}{c}{\scalebox{0.7}{\input{figure_scripts/fig_example_forgetting_legend}}} \\
    \end{tabular}
    \caption{Additional experiments reveal similar trends as seen in Figure \ref{fig:example_forgetting}. These results highlight subgroups with least representation being the most vulnerable to changing predictions.}
	\label{fig:app_example_forgetting_hyper}
\end{figure*}

\newpage
\subsection{Manipulating Group Level Accuracy with Data Order}

We provide additional results in Fig. \ref{fig:app_manipulate_hyper}, highlighting the predictability of model fairness based on the data order. Note that the hyperparameter setting for an additional epoch of fine-tuning during group accuracy manipulation is the same as the setup used for training that particular model. For example, when manipulating models which were trained with batch size $16$, the single epoch of fine-tuning is also done with batch size $16$.

\begin{figure*}[htbp]
	\centering
    \begin{tabular}{c}
    \textbf{Batch Size = 16} \\
    \scalebox{0.9}{\input{figure_scripts/appendix/fig_manipulate_bs_16}} \\
    \textbf{Learning Rate = 0.01} \\
    \scalebox{0.9}{\input{figure_scripts/appendix/fig_manipulate_lr_01}} \\
    \textbf{Model Architecture = \{2048, 64\}} \\
    \scalebox{0.9}{\input{figure_scripts/appendix/fig_manipulate_arch_2048_64}} \\
    \end{tabular}
    \caption{Additional experiments reveal similar trends as seen in Figure \ref{fig:manipulate}. In only a single epoch of training, we are able to manipulate the group level accuracy trade-off, with relatively small impact on overall accuracy.
    }
	\label{fig:app_manipulate_hyper}
\end{figure*}

\begin{figure*}[htbp]
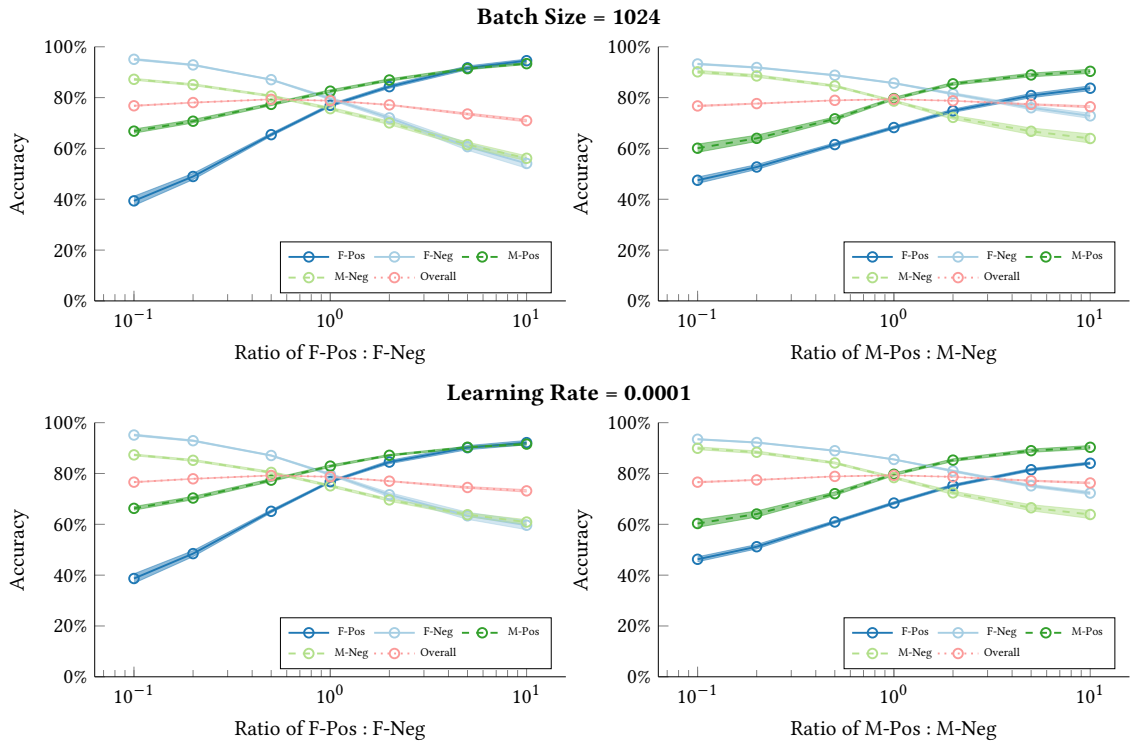

	\centering
    \begin{tabular}{c}
    \textbf{Batch Size = 1024} \\
    \scalebox{0.9}{\input{figure_scripts/appendix/fig_manipulate_bs_1024}} \\
    \textbf{Learning Rate = 0.0001} \\
    \scalebox{0.9}{\input{figure_scripts/appendix/fig_manipulate_lr_0001}} \\
    \end{tabular}
    \caption{Additional experiments as seen in Figure \ref{fig:app_manipulate_hyper}, but for models that did not converge as noted earlier. The results are added for completeness.
    }
	\label{fig:app_manipulate_hyper_add}
\end{figure*}

We also provide results for models with high batch size and low learning rate separately in Fig. \ref{fig:app_manipulate_hyper_add}. Again, note that not only are these models not converged, but they are also fine-tuned on the same inefficient hyperparameter settings. Thus, the trends here are not comparable, but added for completeness.

\newpage

%% file: sections/sec_appendix_dropout.tex
\section{Additional Experiments for Dropout Regularization}
\label{sec:app_dropout}

\begin{figure*}[htbp]
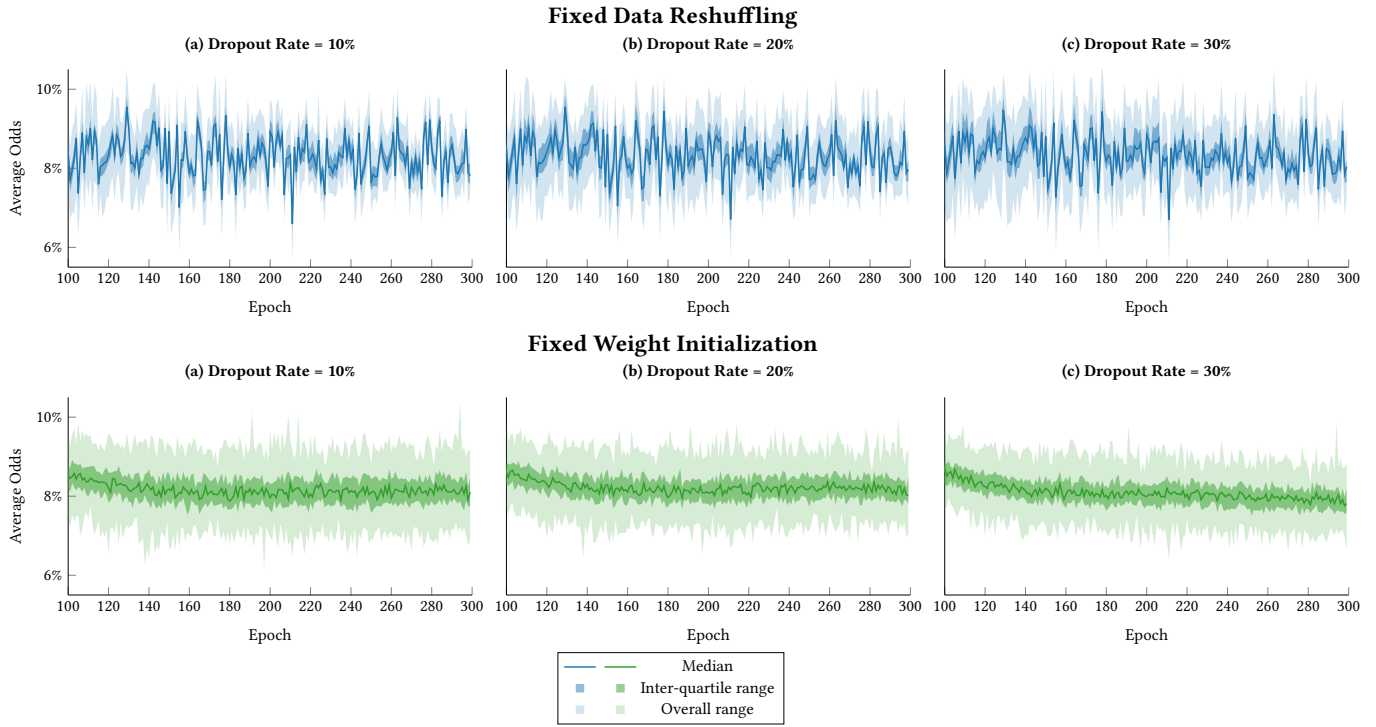

	\centering
        \textbf{Fixed Data Reshuffling}\\
	\scalebox{0.7}{\input{figure_scripts/fig_curve_correlation_dropout_fixed_reshuffling}}\\
        \textbf{Fixed Weight Initialization}\\
	\scalebox{0.7}{\input{figure_scripts/fig_curve_correlation_dropout_fixed_initialization}}
	\scalebox{0.7}{\input{figure_scripts/fig_curve_correlation_legend}}
    \caption{Median, inter-quartile range, and overall range of average odds across 50 different training runs with dropout regularization, while changing only the weight initialization or the random reshuffling, for various dropout rates. The overall range of fairness variance has decreased, and the range of variance across multiple training runs with fixed data reshuffling increases with higher dropout rate, when compared to training without dropout layers. Despite this, the trends of data order dominance over fairness variance are clearly visible even with dropout regularization.
    }
	\label{fig:decouple_correlation_dropout}
\end{figure*}

Another notable hyperparameter in neural model training is dropout regularization \citep{srivastava2014dropout}. Dropout regularization randomly drops a certain percentage (known as dropout rate) of connections between consecutive layers in the model at every forward pass during training. We extend our discussion from Section \ref{sec:variance} to study the impact of dropout on the trends of random reshuffling seen in Figure \ref{fig:decouple_correlation}. More specifically, we repeat the experiments while introducing various rates of dropout in the training setup. The results are collected in Fig. \ref{fig:decouple_correlation_dropout}.

Even with dropout regularization, the impact of data reshuffling on model fairness clearly dominates weight initialization. That is, the trends of high correlation between multiple runs with the same data order, despite starting from different initializations, and lack of any correlation between multiple runs with different data order, even after starting from the same initialization, are still evident even in presence of other sources of randomness.

%% file: sections/sec_appendix_fairness_metrics.tex
\newpage
\section{Additional Experiments for Other Fairness Metrics}
\label{sec:app_fairness}

We repeat the experiment in the main text for two more fairness metrics, Equal Opportunity (EOpp) and Demographic Parity (DP) \cite{hardt2016equality}

Using the same notations as in Section \ref{sec:problem}, EOpp can be defined as,
\begin{align}
    EOpp(f,\Data) &:= |\frac{\sum_{\Data^e} \Indicator_{[f(x)=1 \wedge y=1 \wedge a=0]}}{\sum_{\Data^e} \Indicator_{[y=1 \wedge a=0]}} - \frac{\sum_{\Data^e} \Indicator_{[f(x)=1 \wedge y=1 \wedge a=1]}}{\sum_{\Data^e} \Indicator_{[y=1 \wedge a=1]}}|.
\end{align}

Similarly, DP can be defined as,
\begin{align}
    DP(f,\Data) &:= 1 - min(\frac{\sum_{\Data^e} \Indicator_{[f(x)=1 \wedge a=0]}}{\sum_{\Data^e} \Indicator_{[f(x)=1 \wedge a=1]}}, \frac{\sum_{\Data^e} \Indicator_{[f(x)=1 \wedge a=1]}}{\sum_{\Data^e} \Indicator_{[f(x)=1 \wedge a=0]}}).
\end{align}

\subsection{High Variance in Fairness Scores}

\begin{figure*}[hbtp]
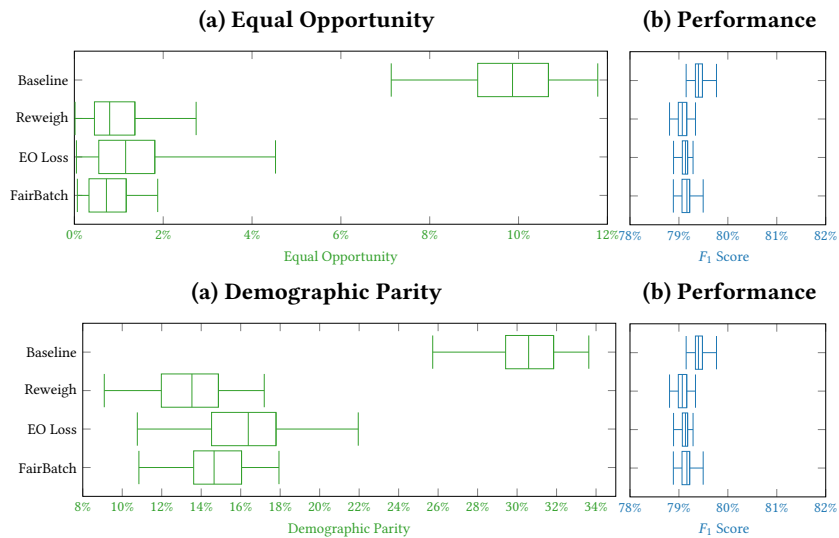

	\centering
	\addtolength{\tabcolsep}{-8pt}
	\begin{tabular}{cc}
	\textbf{(a) Equal Opportunity} & \textbf{(b) Performance} \\[0.2cm]
	\scalebox{0.7}{\input{figure_scripts/fig_variance_introduction_eqopp}} & \scalebox{0.7}{\input{figure_scripts/fig_variance_introduction_fscore}} \\
	\textbf{(a) Demographic Parity} & \textbf{(b) Performance} \\[0.2cm]
	\scalebox{0.7}{\input{figure_scripts/fig_variance_introduction_disimp}} & \scalebox{0.7}{\input{figure_scripts/fig_variance_introduction_fscore}} \\
	\end{tabular}
    \caption{Additional experiments with different fairness metrics for setting in Figure \ref{fig:variance_introduction}. Fairness has a high variance across multiple runs. Note that the x-axis across fairness and F1 score is not similarly scaled for demographic parity to keep the results readable.}
	\label{fig:variance_introduction_metric}
\end{figure*}

We start by repeating the experiment comparing various state-of-the-art bias mitigation techniques and intersecting range of fairness scores in Fig. \ref{fig:variance_introduction_metric}.

\subsection{Weight Initialization and Random Reshuffling}

\begin{figure*}[htbp]
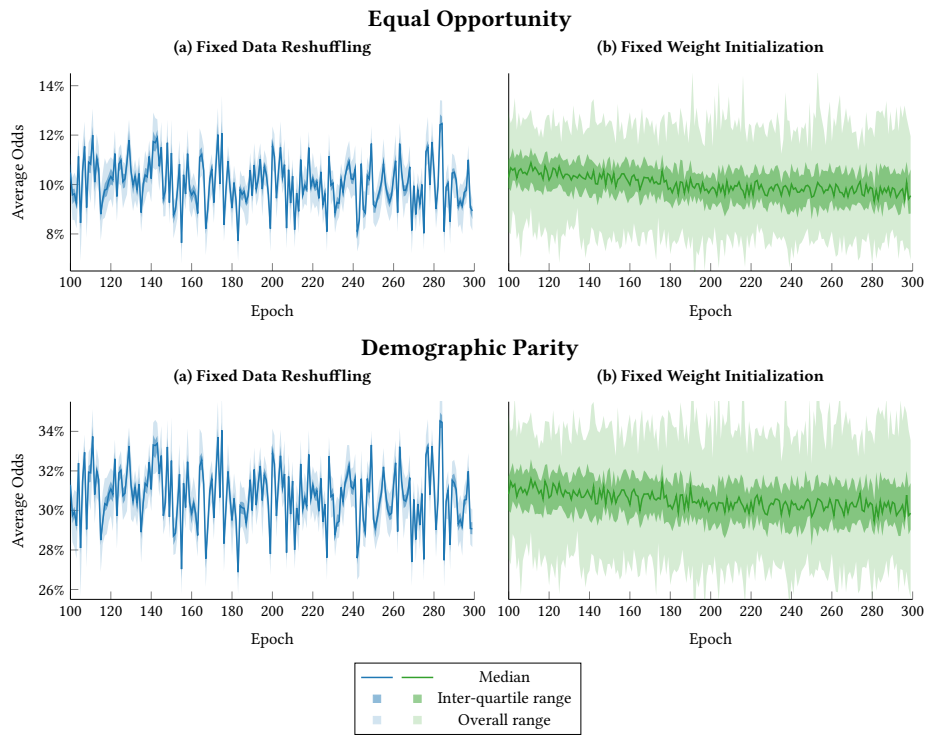

	\centering
	\begin{tabular}{c}
	\textbf{Equal Opportunity} \\
	\scalebox{0.7}{\input{figure_scripts/appendix/fig_curve_correlation_eqopp}} \\
	\textbf{Demographic Parity} \\
	\scalebox{0.7}{\input{figure_scripts/appendix/fig_curve_correlation_disimp}} \\
	\scalebox{0.7}{\input{figure_scripts/fig_curve_correlation_legend}} \\
	\end{tabular}
    \caption{Additional experiments with fairness metric EOpp and DP reveal similar trends as in Figure \ref{fig:decouple_correlation}. These results further highlight the dominant impact of random reshuffling on fairness.}
	\label{fig:app_decouple_correlation_metric}
\end{figure*}

We provide additional results for EOpp and DP for the experiment conducted in Fig. \ref{fig:decouple_correlation} to show correlation between multiple runs on ACSIncome dataset. The results are collected in Fig. \ref{fig:app_decouple_correlation_metric}, and show similar trends as seen in the main text.

\newpage
\subsection{Capturing Variance in a Single Training Run}

\begin{figure*}[hbtp]
    \centering
    \begin{tabular}{cc}
        \multicolumn{2}{c}{\textbf{ACSIncome}} \\
        Equal Opportunity & Demographic Parity \\
        \includegraphics[width=0.4\linewidth]{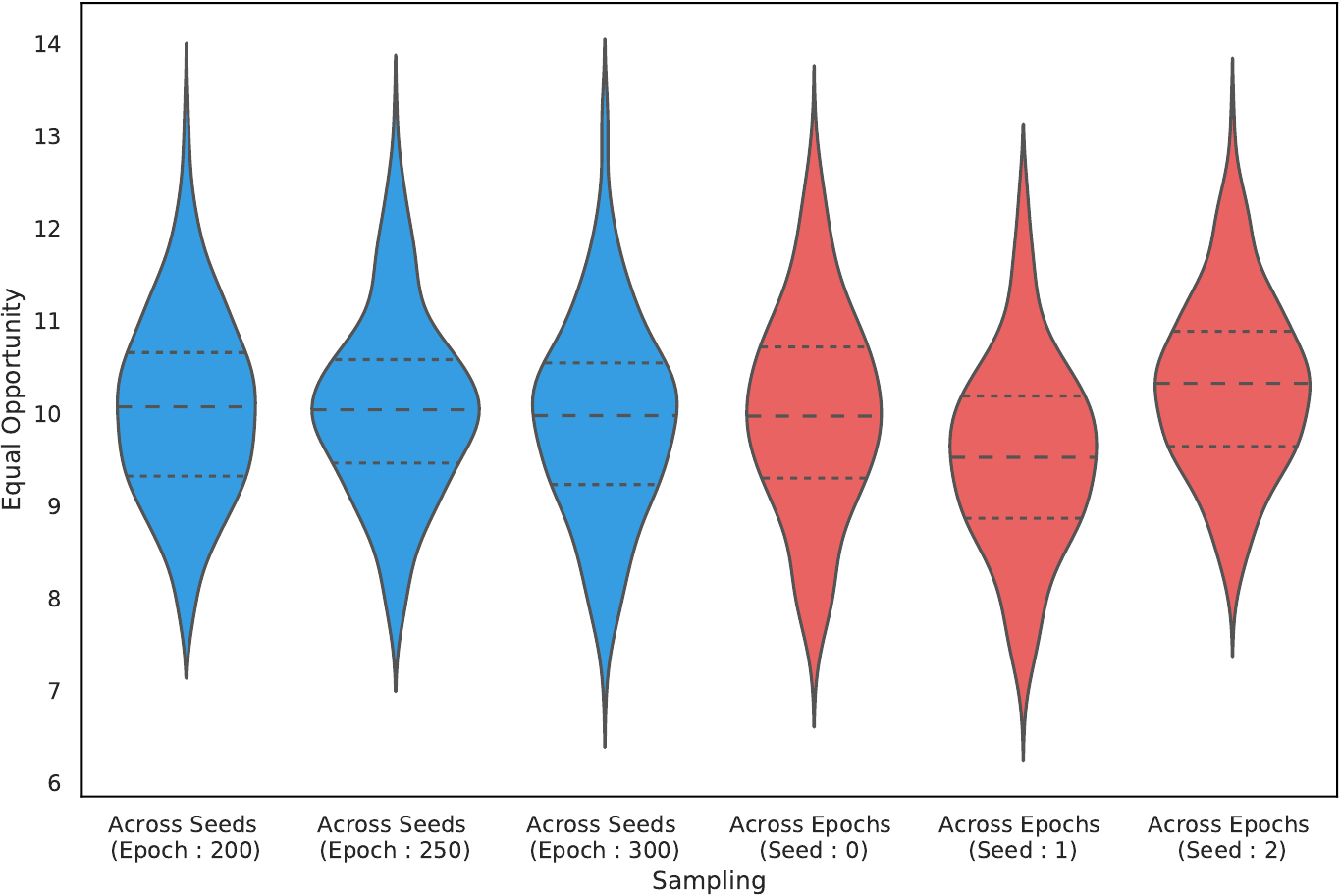} & \includegraphics[width=0.4\linewidth]{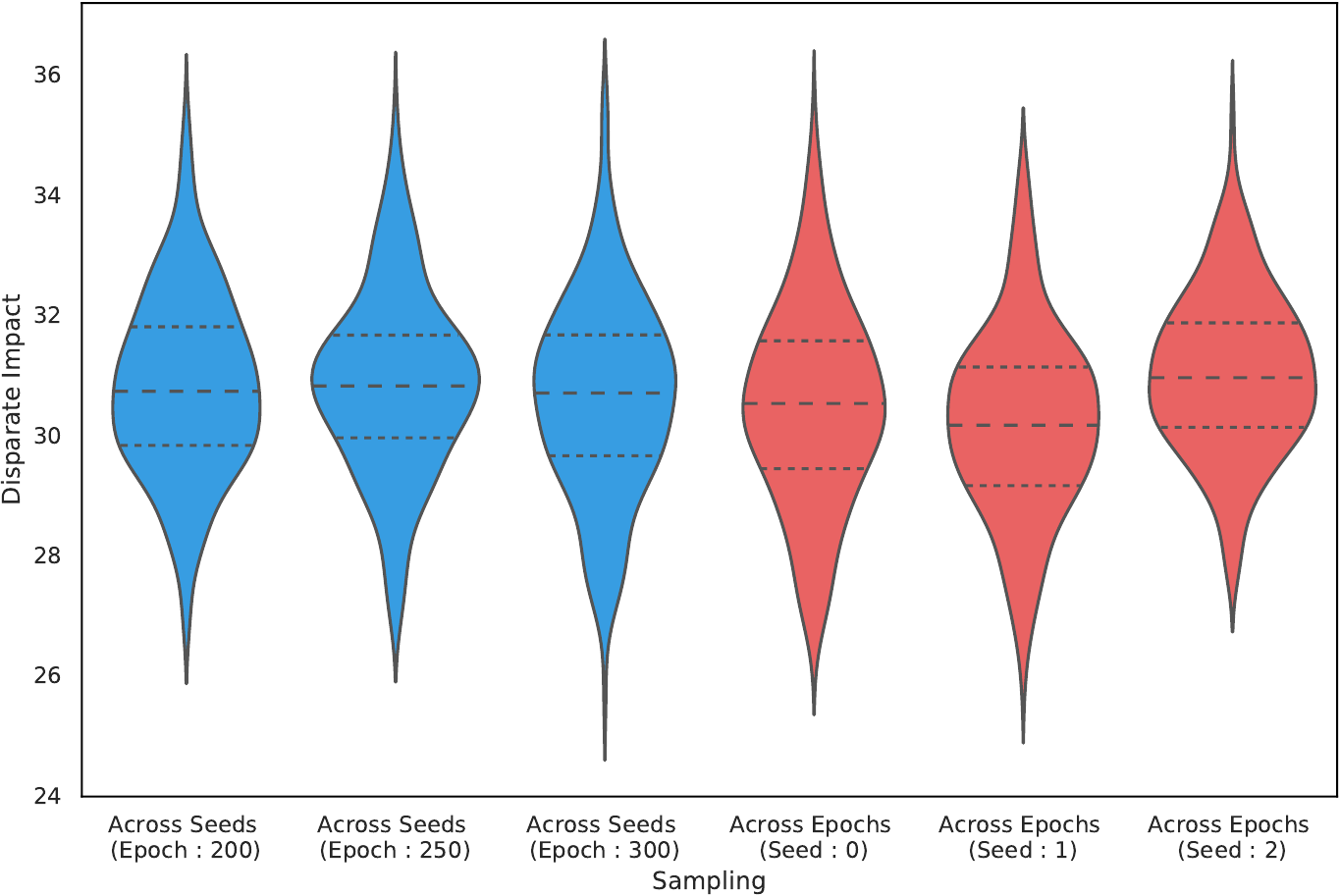} \\
        \multicolumn{2}{c}{\textbf{ACSEmployment}} \\
        Equal Opportunity & Demographic Parity \\
        \includegraphics[width=0.4\linewidth]{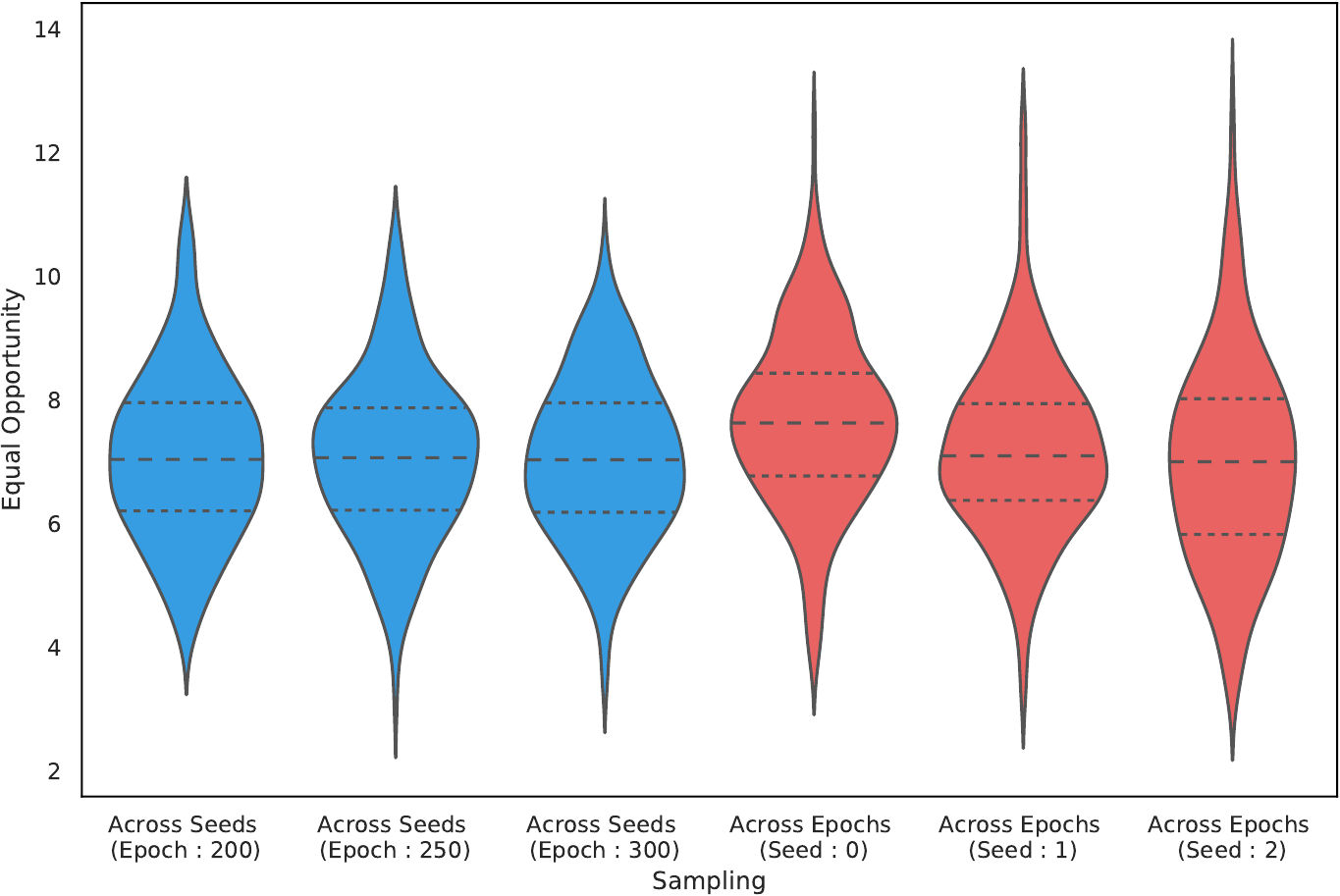} & \includegraphics[width=0.4\linewidth]{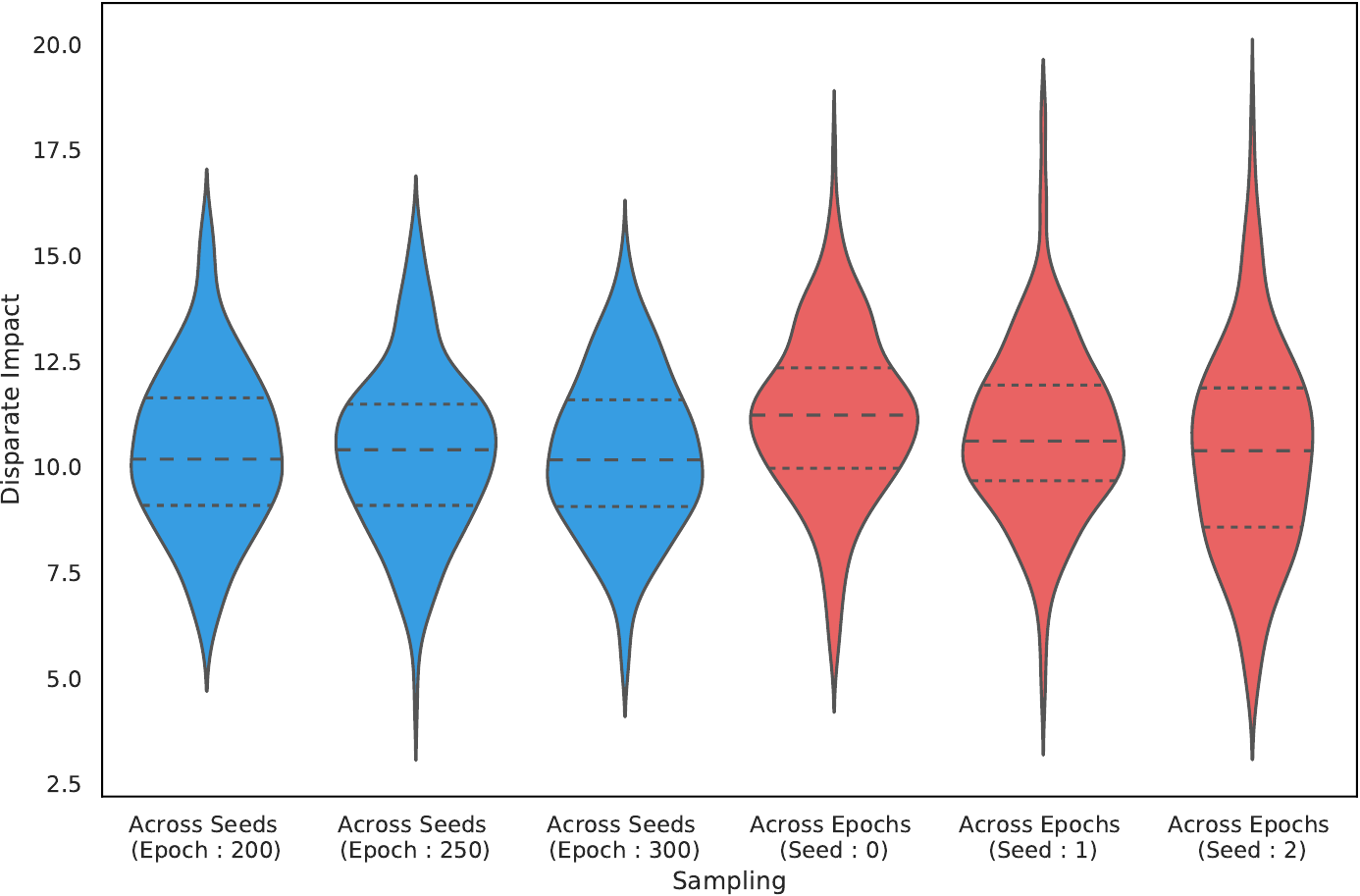} \\
        \multicolumn{2}{c}{\textbf{CelebA}} \\
        Equal Opportunity & Demographic Parity \\
        \includegraphics[width=0.4\linewidth]{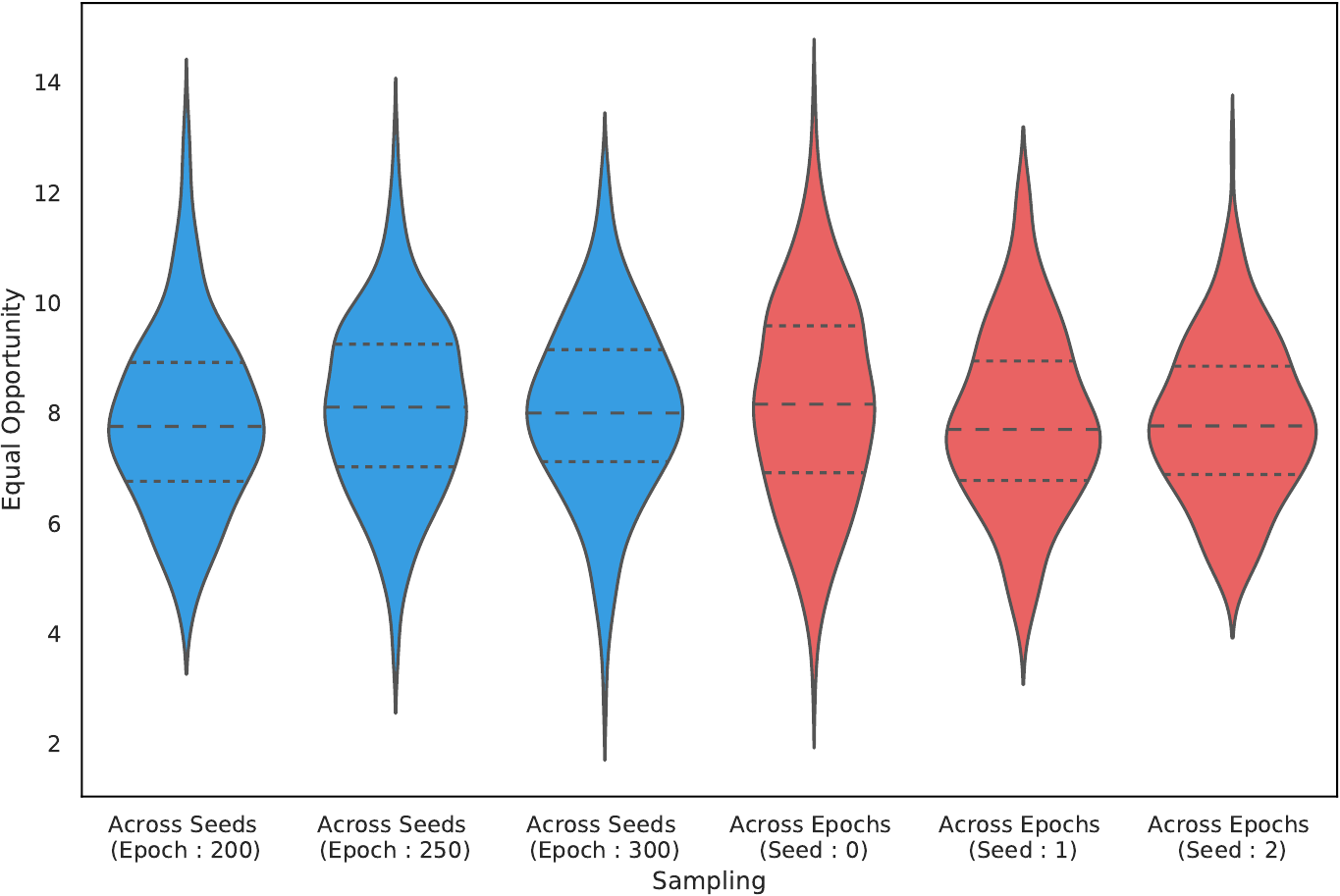} & \includegraphics[width=0.4\linewidth]{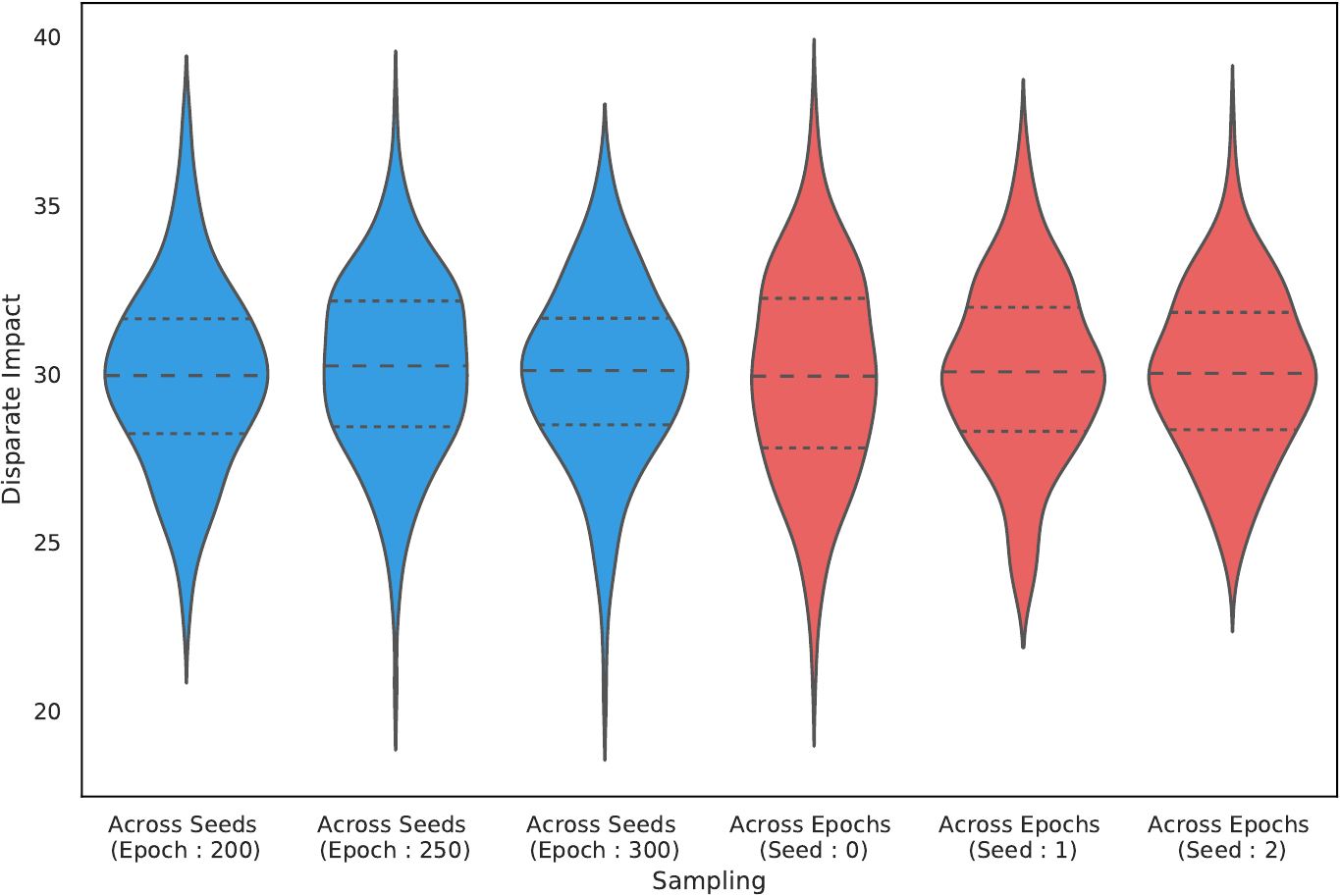} \\
    \end{tabular}
    \caption{Additional experiments for fairness metrics EOpp and DP. Fairness scores across multiple training runs and across epochs in a single training run have similar empirical distributions. Thus, studying this distribution across epochs provides a highly efficient alternative.}
    \label{fig:app_variance_single_run_metric}
\end{figure*}

We show the empirical similarity of distribution across multiple training runs and multiple epochs in a single training run for fairness measures EOpp and DP in Figure \ref{fig:app_variance_single_run_metric}.

%% file: sections/sec_appendix_raw_runs.tex
\newpage
\section{10 Raw Training Runs from Fig. \ref{fig:decouple_correlation}}
\label{sec:app_raw_runs}

\begin{figure*}[htbp]
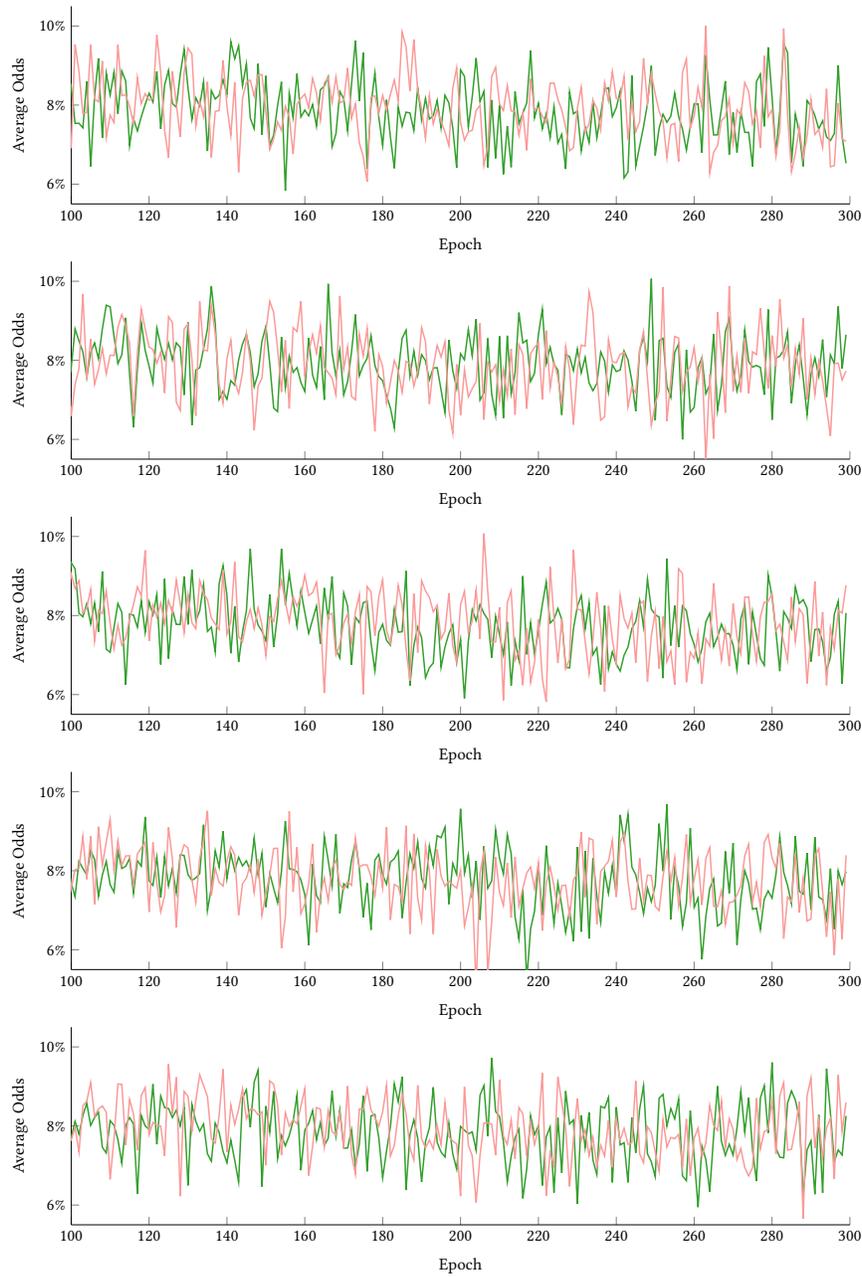

	\centering
	\scalebox{0.7}{\input{figure_scripts/individual_runs_shuffle/fig_run_0-1}}
	\scalebox{0.7}{\input{figure_scripts/individual_runs_shuffle/fig_run_2-3}}
	\scalebox{0.7}{\input{figure_scripts/individual_runs_shuffle/fig_run_4-5}}
	\scalebox{0.7}{\input{figure_scripts/individual_runs_shuffle/fig_run_6-7}}
	\scalebox{0.7}{\input{figure_scripts/individual_runs_shuffle/fig_run_8-9}}
    \caption{10 randomly chosen raw training runs plotted in groups of 2 for fixed weight initialization, but changing random reshuffling.}
	\label{fig:all_runs_shuffle}
\end{figure*}

\begin{figure*}[htbp]
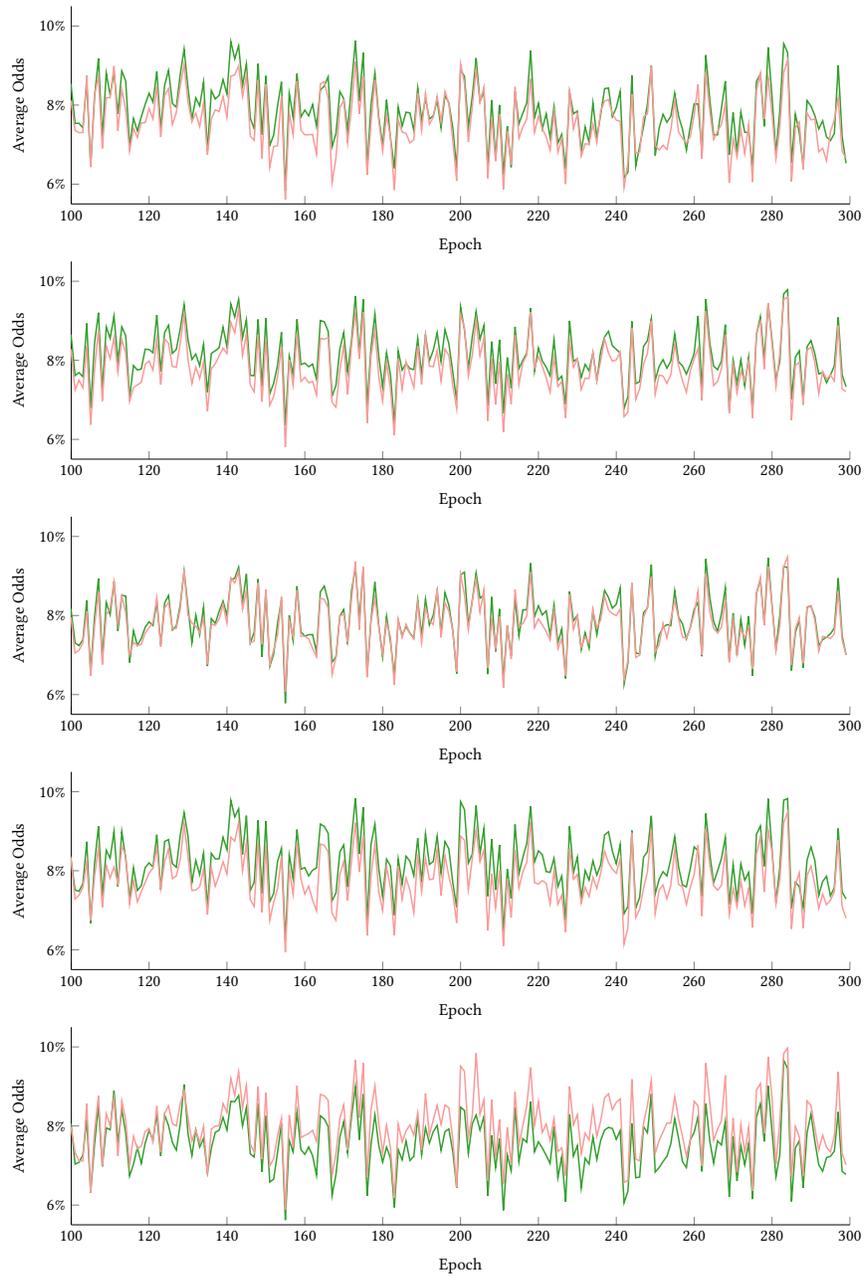

	\centering
	\scalebox{0.7}{\input{figure_scripts/individual_runs_weight/fig_run_0-1}}
	\scalebox{0.7}{\input{figure_scripts/individual_runs_weight/fig_run_2-3}}
	\scalebox{0.7}{\input{figure_scripts/individual_runs_weight/fig_run_4-5}}
	\scalebox{0.7}{\input{figure_scripts/individual_runs_weight/fig_run_6-7}}
	\scalebox{0.7}{\input{figure_scripts/individual_runs_weight/fig_run_8-9}}
    \caption{10 randomly chosen raw training runs plotted in groups of 2 for fixed random reshuffling, but changing weight initialization.}
	\label{fig:all_runs_weight}
\end{figure*}

We plot 10 randomly chosen training runs each for fixed weight initialization and fixed random reshuffling (see Fig. \ref{fig:decouple_correlation}) in Fig. \ref{fig:all_runs_shuffle} and Fig. \ref{fig:all_runs_weight}, respectively. As expected, each individual training run in both settings has high variance across epochs even after convergence. More importantly, the trends of fairness matches closely across multiple runs for fixed random reshuffling, even though they started from different weight initialization.

